\documentclass[pdflatex,sn-vancouver,Numbered]{sn-jnl}

\usepackage{amsmath,amsfonts,amssymb,amsthm}
\usepackage{mathrsfs}
\usepackage{amscd}
\usepackage{graphicx}
\usepackage{xcolor}
\usepackage{booktabs}
\usepackage{array}
\usepackage{tabularx}
\usepackage{multirow}
\usepackage{rotating}
\usepackage{float}
\usepackage{bm}
\usepackage{comment}
\usepackage{makecell}
\usepackage{algorithm}
\usepackage{manyfoot}
\usepackage[title]{appendix}
\usepackage{hyperref}
\hypersetup{hypertexnames=false,
  pdftitle={JDCNet Uses Confidence Gated Privileged Modality Distillation for Cost Preserving X Ray Inference},
  pdfauthor={Bo Ma, Wei Qi Yan, Jinsong Wu, Kun Liu}}
\usepackage{tikz}
\usetikzlibrary{arrows.meta,positioning,shapes.geometric,fit,backgrounds}

\begin{document}

\title[JDCNet Confidence Gated Distillation for X Ray Inference]{JDCNet Uses Confidence Gated Privileged Modality Distillation for Cost Preserving X Ray Inference}

\author*[1]{\fnm{Bo} \sur{Ma}}\email{rcn4743@aut.ac.nz}
\author[1]{\fnm{Wei Qi} \sur{Yan}}
\author[2]{\fnm{Jinsong} \sur{Wu}}
\author[3]{\fnm{Kun} \sur{Liu}}

\affil*[1]{\orgname{Auckland University of Technology}, \orgaddress{\city{Auckland}, \postcode{1024}, \country{New Zealand}}}
\affil[2]{\orgname{Guilin University of Electronic Technology}, \orgaddress{\city{Guilin}, \country{China}}}
\affil[3]{\orgdiv{School of Artificial Intelligence and Data Science}, \orgname{Hebei University of Technology}, \orgaddress{\city{Tianjin}, \country{China}}}

\abstract{Training-only privileged-modality learning uses an expensive modality during training while keeping a single-modality deployment path. We study chest radiograph (X-ray) classification with same-patient computed tomography (CT) available only during training. JDCNet adds an auxiliary CT-teacher target, the teacher's argmax label or its softmax distribution through a Kullback--Leibler term (soft-KL), only on samples whose teacher confidence exceeds a threshold; the deployed model has the same ResNet-18 computation graph as the supervised baseline. On 510 same-patient pairs from the public BIMCV-COVID19 releases of the Valencian Region Medical ImageBank (113 COVID-19, 397 non-COVID), with patient-level five-fold cross-validation and three seeds, 15 of 16 adaptively explored JDCNet configurations improved mean balanced accuracy (BA) over the supervised baseline (average $+0.020$), but none remained significant after Holm adjustment. In a pre-registered experiment with 240 batch-matched supervised restarts on the same splits, batch size had no detectable effect on the baseline (BA $0.620$ versus $0.625$), and when both methods selected one of 16 runs per cell on the validation partition, JDCNet exceeded supervised training by $+0.037$ BA (95\% confidence interval $+0.016$ to $+0.058$; patient-level $+0.007$ to $+0.068$). Logit distillation and contrastive alignment did not help. Teacher diagnostics indicate that the gate acts mainly as CT-informed sample weighting. All models, including JDCNet, performed at chance on an external radiograph set. Confidence-gated CT supervision thus gives a small within-cohort advantage on one public cohort and a binary task; generalization requires independent paired replication.}


\maketitle

\section{Introduction}
\label{sec:introduction}
Training-only privileged-modality learning uses a richer auxiliary modality during training to supervise a lighter modality whose model is the one actually deployed. The deployed model carries the inference cost (latency, memory, energy, bandwidth), so any auxiliary modality that is unavailable at test time can only act through the deployed model's weights. The pattern is attractive whenever the auxiliary modality is expensive to acquire, transmit, or process; video-to-image, depth-to-color-image, multi-view-to-single-view, and high-to-low-resolution settings all share this constraint. Whether the auxiliary signal can be transferred without an inference-time penalty is, however, an empirical question that depends on the cohort, the transfer mechanism, and the evaluation protocol. This paper examines that question for chest X-ray classification with same-patient computed tomography (CT) available only at training time, and treats it as an evaluation problem rather than as a claim of clinical readiness.

Chest X-ray and CT expose this trade-off sharply. Chest X-ray is inexpensive, portable, and widely available, whereas CT provides richer anatomical detail at higher acquisition cost and is often unavailable at the point of care. The question is therefore whether training-time access to CT can improve an X-ray-only model without changing what is computed at inference. Small paired medical cohorts make this question difficult to answer reliably, because split accidents, configuration selection, and source artifacts can all masquerade as method gains.

\subsection{Motivation}

The training-only pattern is most relevant where chest X-ray is the frontline modality and CT is acquired for a subset of patients, for example in emergency departments, in clinics with portable X-ray but no on-site CT, and in screening programs that reserve CT for equivocal cases. In such settings, CT already acquired during routine work-up is a training-time asset even though it cannot be assumed for future patients. A method that uses this asset without adding CT acquisition or computation to the deployed model would keep the deployed X-ray inference graph unchanged. Whether such a method actually improves the X-ray model is the empirical question studied here.

Thoracic imaging models can learn disease representations from X-ray or CT alone~\cite{gao2021covid}, but it is less clear whether CT can act as training-only supervision for an X-ray model. Multimodal fusion typically assumes that both modalities are available at inference~\cite{saeedi2026depression}, which leads to a heavier deployment path. We instead ask a narrower question: can a CT teacher improve an X-ray student \emph{without} requiring CT at test time and \emph{without} changing the deployed student's architecture? Cross-modal transfer is harder than same-modality compression because teacher and student do not share appearance statistics or failure modes, and medical-imaging gains are sensitive to source mismatch and shortcut structure~\cite{raghu2019transfusion,zech2018variable,degrave2021shortcut,ongly2024shortcut}.

Same-patient paired cohorts are small relative to single-modality chest-radiograph corpora because both modalities must exist for the same patient. In this regime the gap between a genuine transfer effect and a sampling or selection artifact is narrow. We therefore treat the evaluation protocol as part of the contribution: the comparison is made on balanced accuracy, the evaluation unit is the patient, every method compared on the 510-patient cohort is evaluated on identical held-out patients within each fold and seed, and we report the complete configuration search, its chronology, and multiplicity-adjusted inference rather than only the best configuration.

In summary, the prediction task is binary COVID-19 versus non-COVID chest X-ray classification; same-patient CT is available only during training; the deployed model receives X-ray only; and the main evidence comes from patient-level five-fold cross-validation on a public 510-patient paired cohort assembled from the BIMCV-COVID19 releases. The paper does not claim clinical readiness.

\subsection{Contributions}

\begin{itemize}
    \item \textbf{Training-only privileged-modality formulation.} We frame CT-to-X-ray transfer as a problem in which CT may shape only the training objective and the deployed model must remain an X-ray-only classifier with the same computation graph as its supervised counterpart.
    \item \textbf{JDCNet confidence-gated auxiliary supervision.} JDCNet applies a CT-teacher auxiliary target only when the teacher is confident, using either the hard argmax label or a soft-KL target, while keeping the class-weighted supervised X-ray loss on every sample. We show that, for a two-class teacher, global temperature scaling of the gate is equivalent to a change of confidence threshold, and we bound the information that the hard target discards.
    \item \textbf{Transparent patient-level evaluation with selection-adjusted inference.} On 510 same-patient paired patients we report all 16 JDCNet configurations with the order in which they were explored, directly computed and multiplicity-adjusted $p$-values, and unselected averages. A pre-registered experiment with 240 batch-matched supervised restarts tests the training-budget difference and gives the baseline the same number of attempts; under this nested comparison JDCNet retains an advantage of $+0.037$ balanced accuracy (95\% CI $[+0.016,+0.058]$ over fold/seed cells and $[+0.007,+0.068]$ over patients), although no individual configuration survives multiplicity adjustment.
    \item \textbf{Mechanism, comparator, and external evidence.} Under the same splits, logit distillation and contrastive alignment show no gain. Teacher diagnostics show that the gate admits most training samples with targets that agree with the ground truth, so the auxiliary loss acts mainly as CT-informed sample weighting. All students, including JDCNet, perform at chance on an external radiograph set, which bounds the claim to the development cohort.
\end{itemize}

\section{Related Work}

\subsection{Efficient Visual Inference under Single-Modality Deployment Constraints}

Efficient visual recognition reduces the parameters, multiply-accumulate operations (MACs), latency, and memory of deployed models. Lightweight backbones such as MobileNetV2~\cite{sandler2018mobilenetv2} and EfficientNet~\cite{tan2019efficientnet} set the deployable operating point, and knowledge distillation (KD)~\cite{hinton2015distilling,romero2014fitnets} trains a small student against a heavier teacher so that teacher-side information is carried at student-side cost; recent chest X-ray work distills large teachers into on-device students~\cite{termritthikun2024explainable}. In the cross-modal version studied here the teacher consumes a modality that the deployed system never sees, so the deployed cost is set entirely by the student. We use this distinction to separate transfer mechanisms that leave the deployed graph unchanged (logit-level and auxiliary-target supervision) from those that add deployable modules (Section~\ref{sec:deployment_efficiency}).

\subsection{Medical Image Classification on Chest X-ray and CT}

Deep learning is a standard tool for thoracic classification. Chest-radiograph benchmarks such as ChestX-ray14/CheXNet, CheXpert, and MIMIC-CXR established the large-cohort context~\cite{rajpurkar2017chexnet,irvin2019chexpert,johnson2019mimiccxr}, and recent work addresses long-tailed multi-label disease classification~\cite{holste2024cxrlt} and label-efficient multi-pathology detection~\cite{tiu2022expert}. Transfer-learning analyses show that apparent gains depend on source--target similarity~\cite{raghu2019transfusion}. COVID-specific studies showed that both chest X-ray and CT support automated classification, but most train and evaluate one modality at a time rather than studying training-only cross-modal transfer~\cite{gao2021covid}, and a prospective multi-hospital evaluation of a COVID-19 chest-radiograph model found substantially lower performance than in development~\cite{sun2022performance}.

\subsection{Transformers and Attention Mechanisms in Medical Imaging}

Transformer-based vision models~\cite{dosovitskiy2021vit,liu2021swin} model long-range spatial dependencies and have been applied to thoracic classification~\cite{gao2021covid}, and hybrid attention designs fuse convolutional features with auxiliary cues at moderate parameter counts~\cite{nemati2025ferha}. Two observations matter here. First, attention-based aggregation motivated the attention-transfer comparator that we evaluated in an early, small-cohort stage of this study. Second, larger attention-based backbones raise a capacity-versus-supervision question: an early feasibility scaffold used lightweight convolutional neural networks (CNNs) at $128{\times}128$, the main 510-patient experiments use ImageNet-pretrained ResNet-18 at $224{\times}224$, and a BiomedCLIP vision-transformer (ViT-B/16) student was evaluated on a smaller balanced BIMCV subset as a capacity control rather than as a deployment candidate.

\subsection{Knowledge Distillation in Medical Imaging}

Knowledge distillation transfers predictive structure from a stronger teacher to a lighter student~\cite{hinton2015distilling,romero2014fitnets}. In medical imaging it has been applied within a single modality for chest X-ray classification, gastric X-ray learning, and hyperspectral image analysis~\cite{ho2020utilizing,li2020soft,sonsbeek2021variational,yue2021self,termritthikun2024explainable}, and recent work frames distillation as image--text representation alignment~\cite{rui2024multimodal}. These studies show that softened teacher predictions can improve compact models, but they mostly operate within one imaging modality. Multimodal foundation models such as BiomedCLIP~\cite{zhang2023biomedclip} provide pretrained medical representations but do not address patient-level classification under training-only CT supervision. Our setting differs in three jointly enforced constraints: teacher and student consume different imaging modalities, the auxiliary modality is available only during training, and the deployed graph must remain an unchanged single-modality classifier.

\subsection{Privileged Information and Training-Only Auxiliary Modalities}

Learning using privileged information studies signals that are available during training but absent at inference~\cite{vapnik2009lupi,lopezpaz2016privileged}. Modality hallucination shows that side information can improve a single-modality model when the auxiliary modality is missing at test time~\cite{hoffman2016hallucination}. In medical imaging, a multimodal teacher has been used to transfer privileged knowledge to a unimodal segmentation student with a distillation loss that down-weights incorrect teacher predictions~\cite{chen2022privileged}, and cross-modal distillation has been used to handle missing modalities at test time~\cite{wang2023learnable}. These works are the closest precedents to the present setting, which can be read as a privileged-modality classification problem: CT is potentially informative during training, but the deployed classifier must remain X-ray only.

\subsection{Cross-Modal Distillation, Teacher Reliability, and Temperature}

Supervision transfer across modalities has been explored through intermediate representation alignment between paired sources~\cite{gupta2016crossmodal}. Generic mechanisms relevant to our comparators include hint-based feature transfer~\cite{romero2014fitnets}, attention transfer~\cite{zagoruyko2017attention}, contrastive representation distillation~\cite{tian2020contrastive}, and multi-scale feature-pyramid aggregation~\cite{lin2017fpn}. Recent cross-modal work shows that paired-modality alignment~\cite{sarkar2022xkd} and disentangling modality-shared from modality-specific signal~\cite{ferrod2025crodinokd} matter. Several methods modulate the teacher signal by its reliability: mixture-of-teacher routing~\cite{li2025mstdistill}, reliability-aware distillation~\cite{wu2023krd}, confidence-aware multi-teacher weighting~\cite{zhang2022confidence}, and calibration-balanced divergences~\cite{amara2022bdkd}. Decoupled KD separates target-class from non-target-class knowledge~\cite{zhao2022decoupled}; for a two-class problem the non-target term carries no information, which is relevant to our hard-versus-soft comparison. The distillation temperature is itself a design axis~\cite{li2023curriculum}, and post-hoc temperature scaling is the standard single-parameter calibration method for neural classifiers~\cite{guo2017calibration}, including class-imbalanced medical classifiers~\cite{rajaraman2022calibration}. Because JDCNet gates on teacher confidence, calibration and temperature are directly relevant (Section~\ref{sec:theoretical_rationale}).

\subsection{Relation to Recent Cross-Modal Privileged-Transfer Work}
\label{sec:relation_recent_prior}

Recent cross-modal medical work includes bridging chest X-ray and CT pulmonary angiography with latent-diffusion priors~\cite{cahan2025cxrctpa}, disentangling modality-shared and modality-specific representations~\cite{zhang2026dante}, and routing a multi-anatomy teacher mixture~\cite{zeng2026kmat}. None of these targets same-patient CT-to-X-ray binary classification with a deployed graph identical to the supervised baseline, and none reports the full set of tested configurations together with the transfer channels that did not help. Table~\ref{tab:sota_comparison} summarizes these design differences. The comparison concerns published methodological design rather than a re-run on our cohort, because several methods target other modalities, tasks, or domains.

\begin{table}[htbp]
\caption{Methodological comparison of JDCNet with related privileged-modality and confidence-aware distillation methods. ``Fixed-cost deploy.'' denotes whether the deployed model is audited to have the same parameters and MACs as a single-modality baseline; ``Negative-channel reporting'' denotes whether the study reports alternative transfer mechanisms that were tested and did not help.}
\label{tab:sota_comparison}
\centering
\scriptsize
\setlength{\tabcolsep}{2.5pt}
\renewcommand{\arraystretch}{0.92}
\begin{tabularx}{\textwidth}{@{}>{\raggedright\arraybackslash\hspace{0pt}}p{0.14\textwidth}>{\raggedright\arraybackslash\hspace{0pt}}p{0.15\textwidth}>{\raggedright\arraybackslash\hspace{0pt}}p{0.07\textwidth}>{\raggedright\arraybackslash\hspace{0pt}}p{0.11\textwidth}>{\raggedright\arraybackslash\hspace{0pt}}p{0.16\textwidth}>{\raggedright\arraybackslash\hspace{0pt}}p{0.14\textwidth}>{\raggedright\arraybackslash\hspace{0pt}}X@{}}
\toprule
Method & Modality pairing & Same-pt.? & Fixed-cost deploy.? & Gating / selection mechanism & Target granularity & Negative-channel reporting \\
\midrule
JDCNet (ours) & CT (train only) $\to$ X-ray & Yes & Yes, X-ray-only & Hard per-sample confidence mask & Argmax label or untempered two-class distribution & Yes \\
Chen et al.~\cite{chen2022privileged} & Multimodal magnetic resonance imaging (MRI; train) $\to$ unimodal & Yes & Not audited & Down-weighting of incorrect teacher outputs & Segmentation maps & No \\
Cahan et al.~\cite{cahan2025cxrctpa} & Chest X-ray $\leftrightarrow$ CT pulmonary angiography & Yes & Not audited & Latent-diffusion prior & Continuous latent representation & No \\
DANTE~\cite{zhang2026dante} & Cross-modal (disentangled) & Not specified & Not audited & Shared/\allowbreak specific disentanglement & Continuous representation & No \\
K-MaT~\cite{zeng2026kmat} & Multi-anatomy teacher mixture & Not CT--X-ray & Not audited & Instance routing across teachers & Continuous & No \\
KRD~\cite{wu2023krd} & Generic (non-medical) & N/A & N/A & Continuous reliability weight & Softened distribution & No \\
MST-Distill~\cite{li2025mstdistill} & Generic mixture-of-teacher & N/A & N/A & Continuous instance routing & Softened distribution & No \\
BDKD~\cite{amara2022bdkd} & Generic calibration-balanced KD & N/A & N/A & Continuous divergence weighting & Softened distribution & No \\
\bottomrule
\end{tabularx}
\end{table}

\subsection{Evidence Robustness in Small Medical-Imaging Cohorts}

Clinical-impact reviews~\cite{kelly2019clinicalimpact}, COVID-19 prediction-model appraisals~\cite{wynants2020prediction,roberts2021pitfalls}, hidden-stratification analyses~\cite{oakdenrayner2020hidden}, shortcut analyses in chest-radiograph learning~\cite{degrave2021shortcut,zech2018variable,ongly2024shortcut}, subgroup generalization studies~\cite{ahluwalia2023subgroup}, and biomedical benchmarking studies~\cite{maierhein2018rankings,varoquaux2022failures} share a common warning: small or weakly described evaluation cohorts produce optimistic rankings, and internal cross-validation does not establish robustness to unseen distributions~\cite{zhang2023when}. Varoquaux and Cheplygina~\cite{varoquaux2022failures} specifically note that selecting among many configurations on the evaluation data inflates reported performance. This concern is central here because the paired cohort is small, the configuration space was explored adaptively, and COVID-era thoracic models are exposed to source shift. We therefore report the full search, its order, and selection-adjusted inference as part of the results.

\section{Methodology}
\label{sec:methodology}

\subsection{Notation and Abbreviations}
\label{sec:notation}
\textbf{Abbreviations.} AP: anteroposterior; AUC: area under the receiver operating characteristic curve; BA: balanced accuracy, the mean of sensitivity and specificity; BIMCV: Valencian Region Medical ImageBank (Banco de Im\'agenes M\'edicas de la Comunidad Valenciana), whose BIMCV-COVID19+ and BIMCV-COVID19$-$ releases supply our cohort~\cite{delaiglesiavaya2020bimcv}; CE: cross-entropy; CI: confidence interval; CNN: convolutional neural network; CPU: central processing unit; CT: computed tomography; DRR: digitally reconstructed radiograph, a synthetic X-ray-like projection rendered from a CT volume; ECE: expected calibration error; GPU: graphics processing unit; KD: knowledge distillation; KL: Kullback--Leibler divergence; MAC: multiply-accumulate operation; MIDRC: Medical Imaging and Data Resource Center, the source of the external radiograph set; SD: standard deviation; ViT: vision transformer; wCE: class-weighted cross-entropy. Macro-F1 denotes the unweighted mean of the per-class F1 scores.

\textbf{Method names.} \textbf{JDCNet}: the method introduced here, in which a teacher-confidence mask selects the samples on which an auxiliary cross-entropy (hard variant) or KL (soft-KL variant) target is applied. \textbf{Plain logit KD}: cross-modal distillation of temperature-softened teacher logits on every sample. \textbf{Gated logit KD}: logit KD restricted to teacher-correct samples and weighted by teacher confidence (Section~\ref{sec:comparator_objective}). \textbf{CT-only reference}: a CT classifier trained and evaluated on the same patients; it is a training-only reference for how much class signal the CT view carries, not a mathematical upper bound on student performance. \textbf{Module-augmented pilot}: a historical lightweight-CNN variant with three optional operators labeled DPE (a $1{\times}1$ spatial reweighting block), MHRA (a multi-head attention retain-gate block), and DFPN (a three-scale feature-pyramid neck); these are study-specific labels for standard operator combinations, not abbreviations of established methods.

\subsection{Task Formulation}
We study binary COVID-19 versus non-COVID classification with two modalities, chest X-ray and CT. Let $m \in \{\mathrm{xr}, \mathrm{ct}\}$ denote the modality, $x^{(m)}$ an input from modality $m$, and $y \in \{0,1\}$ the label. The objective is to learn an X-ray student $f_S$ while using a CT teacher $f_T$ only during training. The evaluation unit is the patient, and only X-ray is available at test time. This formulation is training-only CT supervision for X-ray deployment, not inference-time multimodal fusion.

\subsection{JDCNet: Confidence-Gated Cross-Modal Distillation}
\label{sec:jdcnet}

\begin{figure}[!t]
\centering
\resizebox{0.98\textwidth}{!}{%
\begin{tikzpicture}[
  box/.style={draw, rounded corners, align=center, minimum height=8mm, minimum width=30mm, font=\small, inner sep=3pt},
  ctbox/.style={box, fill=gray!12, draw=gray!60},
  xrbox/.style={box, fill=blue!8, draw=blue!55},
  gatebox/.style={box, fill=orange!12, draw=orange!70},
  lossbox/.style={box, fill=green!10, draw=green!60},
  gonebox/.style={box, fill=gray!4, draw=gray!45, dashed, text=gray!55},
  predbox/.style={box, fill=green!10, draw=green!60},
  arr/.style={-{Latex[length=2mm]}, thick},
  darr/.style={-{Latex[length=2mm]}, thick, red, dashed}
]
\node[font=\bfseries\small] at (0,6.2) {Training-time graph};
\node[ctbox] (ctimg) at (-2.0,5.1) {CT view $x^{ct}$\\(mid-slice or 3-slice)};
\node[xrbox] (xrimg) at (2.0,5.1) {X-ray $x^{xr}$\\(same patient)};
\node[ctbox] (ctenc) at (-2.0,3.7) {Frozen CT teacher $f_T$\\(trained on this fold)};
\node[xrbox] (xrenc) at (2.0,3.7) {X-ray student $f_S$\\ResNet-18};
\node[gatebox] (gate) at (-2.0,2.3) {Confidence gate\\$\mathcal{M}=\{i: c_i>\tau_{\mathrm{gate}}\}$};
\node[gatebox] (target) at (-2.0,0.9) {Auxiliary target on $\mathcal{M}$: argmax\\(hard) or $\sigma(f_T)$ (soft-KL)};
\node[lossbox] (loss) at (0.6,-0.6) {$\mathcal{L}=\mathrm{wCE}(f_S(x^{xr}),y)+\lambda\,\mathcal{L}_{\text{aux}}$};
\draw[arr] (ctimg) -- (ctenc);
\draw[arr] (xrimg) -- (xrenc);
\draw[arr] (ctenc) -- (gate);
\draw[arr] (gate) -- (target);
\draw[arr] (target) |- (loss);
\draw[arr] (xrenc) -- (xrenc |- loss.north);
\draw[darr] (loss.east) -| ++(1.4,0) |- (xrenc.east);
\draw[gray!50, thick, dashed] (5.3,-1.2) -- (5.3,6.4);
\node[font=\bfseries\small] at (9.0,6.2) {Inference-time graph};
\node[gonebox] (ctimg2) at (7.0,5.1) {CT input\\\emph{removed}};
\node[xrbox] (xrimg2) at (11.0,5.1) {X-ray $x^{xr}$};
\node[gonebox] (ctenc2) at (7.0,3.7) {CT teacher\\\emph{removed}};
\node[xrbox] (xrenc2) at (11.0,3.7) {X-ray student $f_S$\\(trained weights)};
\node[gonebox] (gate2) at (7.0,2.3) {Gate and auxiliary\\loss \emph{removed}};
\node[predbox] (pred) at (11.0,-0.6) {Prediction $f_S(x^{xr})$};
\draw[arr] (xrimg2) -- (xrenc2);
\draw[arr] (xrenc2) -- (pred);
\end{tikzpicture}%
}
\caption{JDCNet training-time and inference-time computation graphs. \textit{Left:} for each cross-validation fold and seed, a CT teacher trained only on that fold's training patients is frozen; its confidence $c_i=\max_k p^{ct}_{i,k}$ selects the samples on which an auxiliary target (the teacher's argmax label, or its softmax distribution for the soft-KL variant) is added to the class-weighted cross-entropy on the ground-truth label. Gradients flow only into the student (red dashed arrow). \textit{Right:} at deployment the CT input, teacher, gate, and auxiliary loss are removed; only the trained X-ray student remains, so the deployed graph is the same ResNet-18 classifier as the supervised baseline (Table~\ref{tab:efficiency}).}
\label{fig:jdcnet_train_infer_graph}
\end{figure}

JDCNet (Fig.~\ref{fig:jdcnet_train_infer_graph}) uses the CT teacher's prediction as an auxiliary target only on samples that the teacher classifies confidently. Let $p^{ct}_i = \sigma(f_T(x^{(ct)}_i))$ be the teacher's softmax output for patient $i$, $c_i=\max_k p^{ct}_{i,k}$ its confidence, and $\hat{y}^{ct}_i = \arg\max_k p^{ct}_{i,k}$ its predicted label. The per-batch confidence mask is $\mathcal{M} = \{ i : c_i > \tau_{\mathrm{gate}} \}$, and the student loss is
\begin{equation}
\label{eq:jdcnet_loss}
\mathcal{L}_{\text{JDCNet}} = \mathrm{wCE}\!\bigl(f_S(x^{(xr)}),\,y\bigr) + \lambda\;\mathcal{L}_{\text{aux}}\!\bigl(f_S(x^{(xr)})_{\mathcal{M}},\,f_T(x^{(ct)})_{\mathcal{M}}\bigr),
\end{equation}
where $\mathrm{wCE}$ is cross-entropy with inverse-frequency class weights computed on the training partition (the same loss as the supervised baseline), $\lambda$ is the auxiliary weight, and $\mathcal{L}_{\text{aux}}$ is averaged over $\mathcal{M}$ (and is zero when $\mathcal{M}$ is empty). The \textbf{hard} variant uses the teacher's argmax label,
\begin{equation}
\label{eq:aux_hard}
\mathcal{L}_{\text{aux}} = \mathrm{CE}\bigl(f_S(x^{(xr)})_{\mathcal{M}},\,\hat{y}^{ct}_{\mathcal{M}}\bigr),
\end{equation}
and the \textbf{soft-KL} variant uses the teacher distribution,
\begin{equation}
\label{eq:aux_softkl}
\mathcal{L}_{\text{aux}} = T^2 \cdot \mathrm{KL}\Bigl(\sigma\bigl(f_T(x^{(ct)})_{\mathcal{M}}/T\bigr) \,\Big\|\, \sigma\bigl(f_S(x^{(xr)})_{\mathcal{M}}/T\bigr)\Bigr),
\end{equation}
with $T{=}1$ (the untempered softmax) in all reported soft-KL runs. The gate does not use the ground-truth label, so confidently wrong teacher predictions are admitted; the always-on $\mathrm{wCE}$ term keeps the teacher signal auxiliary. The student is an ImageNet-pretrained ResNet-18; at deployment the teacher branch is dropped and the deployed graph is identical to the supervised baseline (Section~\ref{sec:deployment_efficiency}). The governing choices are the CT teacher view (mid-slice or 3-slice), $\tau_{\mathrm{gate}}$, $\lambda$, and the target type. Algorithm~\ref{alg:jdcnet} gives the training procedure.

\begin{algorithm}[t]
\caption{Confidence-gated privileged-modality distillation (JDCNet) within one cross-validation fold and seed. Deployment uses the X-ray student $f_S$ alone.}
\label{alg:jdcnet}
\setlength{\fboxsep}{4pt}
\fbox{\parbox{0.92\columnwidth}{\footnotesize
\textbf{Input:} training and validation partitions of the fold, paired $\{(x^{xr}_i, x^{ct}_i, y_i)\}$; gate $\tau_{\mathrm{gate}}$; weight $\lambda$; target type $\in\{\text{hard},\text{soft-KL}\}$. The test partition is never used for training or selection. \\[2pt]
\textbf{Stage 1.} Train the CT teacher $f_T$ on the training partition $\{(x^{ct}_i, y_i)\}$, select its checkpoint by validation BA, and freeze it. \\[2pt]
\textbf{Stage 2.} For each epoch and minibatch $B$ of the training partition: \\
\quad 1.\ Compute $p^{ct}_i = \sigma(f_T(x^{ct}_i))$ and $c_i = \max_k p^{ct}_{i,k}$ for $i\in B$ (no gradient).\\
\quad 2.\ Form $\mathcal{M} = \{i\in B : c_i > \tau_{\mathrm{gate}}\}$.\\
\quad 3.\ $\mathcal{L}_{ce} \leftarrow \mathrm{wCE}(f_S(x^{xr}_B), y_B)$ on all samples.\\
\quad 4.\ If \emph{hard}: $\mathcal{L}_{aux} \leftarrow \mathrm{CE}(f_S(x^{xr}_{\mathcal{M}}), \arg\max p^{ct}_{\mathcal{M}})$; if \emph{soft-KL}: $\mathcal{L}_{aux} \leftarrow \mathrm{KL}(p^{ct}_{\mathcal{M}} \,\|\, \sigma(f_S(x^{xr}_{\mathcal{M}})))$.\\
\quad 5.\ Update $f_S$ on $\mathcal{L}_{ce} + \lambda \cdot \mathcal{L}_{aux}$.\\
\quad Keep the student checkpoint with the highest validation BA. \\[2pt]
\textbf{Deployment.} Discard $f_T$, $\mathcal{M}$, and the auxiliary loss; evaluate $f_S(x^{xr})$ on the test partition.
}}
\end{algorithm}

\subsection{Design Rationale, Temperature Scaling, and Hard Targets}
\label{sec:theoretical_rationale}

\textbf{Relation to confidence-aware distillation.} Confidence-aware KD designs~\cite{wu2023krd,li2025mstdistill,zhang2022confidence,amara2022bdkd} typically modulate \emph{how strongly} a softened teacher distribution is enforced through a continuous weight. JDCNet combines two different choices: a hard binary mask that either admits or excludes a sample's auxiliary loss, and an auxiliary target that is either the argmax label or the untempered two-class distribution. Our gated logit-KD comparator (Section~\ref{sec:comparator_objective}) differs from JDCNet in several respects at once (distillation temperature $T{=}4$ versus $1$, a teacher-correctness filter that uses the ground-truth label, a continuous confidence weight, the loss mixing coefficient, and the training batch size), so the comparison between them does not isolate the effect of target shape. We therefore state two mechanistic hypotheses, motivated by but not demonstrated by our data; the teacher diagnostics in Section~\ref{sec:jdcnet_results_main} bear on them. \emph{Information-channel hypothesis:} a temperature-softened CT distribution may carry CT-specific structure that has no counterpart in projective X-ray appearance, whereas an argmax label or untempered two-class target retains only the class decision and its coarse confidence. \emph{Bias hypothesis:} a continuous weight attenuates but does not remove the contribution of unreliable teacher predictions, whereas a hard mask removes it entirely when the threshold separates reliable from unreliable predictions. Testing either hypothesis requires a factorial comparison that varies one design factor at a time under a matched training budget.

\textbf{Temperature scaling of the gate is a change of threshold.} Post-hoc temperature scaling divides the logits by a global calibration temperature $T_{\mathrm{g}}>0$ to correct over- or under-confidence~\cite{guo2017calibration}. For a two-class teacher with logit margin $z_i$, the confidence is $c_i=\sigma(|z_i|)$ and the temperature-scaled confidence is $\sigma(|z_i|/T_{\mathrm{g}})$, where for a scalar argument $\sigma(u)=1/(1+e^{-u})$ is the logistic function and $\mathrm{logit}(\tau)=\ln\{\tau/(1-\tau)\}$ is its inverse. Hence
\begin{equation}
\label{eq:temperature_threshold}
\sigma(|z_i|/T_{\mathrm{g}}) > \tau \iff c_i > \tau' = \sigma\bigl(T_{\mathrm{g}}\,\mathrm{logit}(\tau)\bigr),
\end{equation}
and the argmax label is unchanged. For the hard variant, gating on temperature-calibrated confidence is therefore exactly equivalent to gating on raw confidence with a different threshold. For example, $T_{\mathrm{g}}\in\{0.5,1,2\}$ at $\tau=0.70$ corresponds to raw thresholds $\{0.60,0.70,0.84\}$, and at $\tau=0.80$ to $\{0.67,0.80,0.94\}$. Fitting a calibration temperature on validation data would thus re-index $\tau$ rather than add a new safeguard. Temperature scaling also cannot correct the ranking of teacher predictions: a confidently wrong prediction remains confidently wrong (Table~\ref{tab:gate_case_examples}, Case~B). Class-conditional thresholds or vector scaling are not equivalent to a single threshold and remain open. For the soft-KL variant, the distillation temperature $T$ of Eq.~\eqref{eq:aux_softkl} additionally changes the target distribution; all soft-KL runs reported here use $T{=}1$.

\textbf{Information discarded by the hard target.} For two classes, the teacher distribution is fully described by its confidence $c_i$, and the non-target-class term of decoupled KD vanishes~\cite{zhao2022decoupled}. Replacing an admitted teacher distribution by its argmax label discards only the confidence magnitude above the threshold: the divergence between the one-hot target and the teacher distribution is $\mathrm{KL}(\mathbf{e}_{\hat{y}}\|p^{ct}_i)=-\ln c_i<-\ln\tau$, that is, below $0.36$~nats at $\tau=0.70$ and below $0.22$~nats at $\tau=0.80$, and the discarded teacher entropy is below $0.88$ and $0.72$~bits, respectively. Uncertainty below the threshold is not passed to the student by either variant, because those samples are excluded. Empirically (Section~\ref{sec:jdcnet_results_main}) the soft-KL target was on average $+0.009$ BA higher than the hard target at matched teacher view, $\tau$, and $\lambda{=}1.0$, a difference well inside the sampling uncertainty.

\textbf{Extension beyond two classes.} Equation~\eqref{eq:jdcnet_loss} applies unchanged to $K$ mutually exclusive classes, with $c_i=\max_k p^{ct}_{i,k}$; the equivalence in Eq.~\eqref{eq:temperature_threshold} no longer holds exactly for $K>2$, and class-conditional thresholds may be needed for imbalanced label spaces. For multi-label chest radiograph findings~\cite{holste2024cxrlt}, the natural analog is a per-finding sigmoid teacher with a per-finding mask. Neither extension is evaluated here because no public multi-disease same-patient CT--X-ray cohort was available to us.

\textbf{Threshold values.} The analysis plan written before the JDCNet runs listed $\tau_{\mathrm{gate}}\in\{0.60,0.70,0.80\}$ and $\lambda\in\{0.3,0.5,1.0\}$ as candidate values and the choice between hard and soft targets as open. The configurations actually run, and their order, are described in Section~\ref{sec:statistical_protocol}.

\subsection{Comparator Mechanisms}
\label{sec:comparators}
The 510-patient comparator runs use the same patient-level splits and the same ResNet-18 student architecture as JDCNet: plain logit KD and gated logit KD with each of four CT teacher views (mid-slice, 3-slice stack, axial mean projection, and DRR), and a two-stage contrastive-alignment comparator (pretraining with the InfoNCE contrastive loss on patient-paired CT/X-ray batches followed by supervised fine-tuning) with two teacher views and two temperatures. A BiomedCLIP ViT-B/16 student was evaluated only on a 226-patient balanced BIMCV subset. Attention transfer, feature hints, generic KD baselines, and the module-augmented pilot were evaluated only in the early small-cohort stage. These cohort differences are stated wherever the comparators are discussed.

\subsection{Comparator Training Objective}
\label{sec:comparator_objective}
The logit-KD comparators optimize
\begin{equation}
\label{eq:logit_kd_loss}
L = (1-\alpha)\,\mathrm{wCE}\bigl(f_S(x^{(\mathrm{xr})}),y\bigr) + \alpha T^2\,\frac{1}{|B|}\sum_{i\in B} w_i\,\mathrm{KL}\Bigl(\sigma\bigl(f_T(x^{(\mathrm{ct})}_i)/T\bigr)\,\Big\|\,\sigma\bigl(f_S(x^{(\mathrm{xr})}_i)/T\bigr)\Bigr),
\end{equation}
with $T{=}4$ and $\alpha{=}0.6$. Plain logit KD uses $w_i{=}1$. Gated logit KD uses $w_i = c_i\cdot\mathbf{1}[c_i\geq 0.50]\cdot\mathbf{1}[\hat{y}^{ct}_i=y_i]$, that is, a continuous confidence weight restricted to samples that the teacher classifies correctly; for two classes the $0.50$ threshold is not restrictive, so the effective filter is teacher correctness. Gradients flow only through the student.

\subsection{Statistical Analysis}
\label{sec:statistical_protocol}

\textbf{Endpoint and decision rule.} The primary endpoint is the paired difference in test-partition balanced accuracy, $\Delta\mathrm{BA}$, between a method and the supervised X-ray baseline evaluated on the same held-out patients, over five patient-level folds and seeds 42--44 ($n{=}15$ paired fold/seed cells). The 95\% CI is a percentile bootstrap (10{,}000 resamples) over the 15 cells. Before the JDCNet runs, the analysis plan fixed a descriptive decision rule: mean $\Delta\mathrm{BA}\geq+0.03$ and lower CI bound $>0$. The 15 cells are not independent, because the same patients appear in the test folds of all three seeds, so cell-level intervals and tests understate uncertainty about new patients; we therefore also report a patient-level bootstrap (below).

\textbf{Analysis chronology and selection.} The JDCNet configurations were not registered as a fixed grid in advance. Before JDCNet was proposed, the 510-patient cohort had already been used for the CT-only reference and logit-KD comparisons. JDCNet was then explored in three sequential stages on the same cohort and splits. (i)~An initial grid of eight hard-target configurations (teacher view $\in\{$mid, 3-slice$\}$, $\tau_{\mathrm{gate}}\in\{0.70,0.80\}$, $\lambda\in\{0.5,1.0\}$), a subset of the planned ranges: all eight had positive means and none met the decision rule. (ii)~Extension A, added after inspecting stage (i): the same views and thresholds with a hard target and $\lambda{=}1.5$, outside the planned $\lambda$ range. (iii)~Extension B: the same views and thresholds with a soft-KL target and $\lambda{=}1.0$. The two configurations that met the decision rule came from the extension stages. Configuration choice and performance estimation therefore used the same cross-validated test-partition results, and the estimates of the best configurations are subject to selection optimism (the winner's curse)~\cite{varoquaux2022failures}.

\textbf{Directly computed $p$-values and multiplicity.} The per fold/seed test-partition records of all 16 configurations are archived. For each configuration we report the paired $t$-test on the 15 cell differences, with Holm (family-wise error) and Benjamini--Hochberg (false discovery rate) adjustment across the 16 configurations; Appendix Table~\ref{tab:app_initial_grid_tests} adds Wilcoxon signed-rank, exact sign, and bootstrap $p$-values. As unselected summaries we report the per-cell average over all 16 configurations and over the eight initial configurations.

\textbf{Pre-registered supplementary experiment.} To address selection and the training-budget difference described below, we wrote an analysis plan before any new run and then trained, on the original splits and with the original code, 16 supervised restarts per fold/seed cell at the batch size used by JDCNet (128), differing only in the random seed (240 runs). The pre-specified primary analysis is a nested comparison: in every cell, the JDCNet configuration with the highest validation BA among the 16 configurations and the supervised restart with the highest validation BA among the 16 restarts are each evaluated once on the test partition, and their paired difference is summarized over the 15 cells (bootstrap CI, paired $t$-test). The same best-of-16 selection is thus applied to both methods without using test data for selection. Secondary analyses compare each configuration with the first restart (the batch-matched baseline) and with the mean of the 16 restarts of the same cell. Because the evaluation pipeline stored only aggregate metrics, we re-ran inference for every checkpoint and stored per-patient probabilities; the recomputed test BA matched the recorded value for 613 of the 615 runs with recorded test metrics (the other two differed by $0.006$ and $0.022$ BA, the latter one patient), and cell-level statistics use the recorded values. These predictions give a patient-level paired bootstrap (10{,}000 resamples of patients; for each seed, the out-of-fold predictions of the five folds are pooled, and BA is averaged over seeds) and the per-image external analysis.

\textbf{Batch-size difference.} The supervised baseline, the CT-only references, and the logit-KD comparators were trained in one stage with batch size 512, whereas JDCNet and the contrastive comparator used batch size 128. With about 306 training patients per fold, the baseline received one optimizer update per epoch and JDCNet three. The supplementary experiment tests whether this difference explains the JDCNet results (Section~\ref{sec:batch_matched_reruns}).

Table~\ref{tab:main_decision_summary} summarizes the 510-patient comparisons. It separates whether the CT view carries class signal (CT-only reference), whether JDCNet moves the X-ray student, whether the result survives the supplementary controls, and whether alternative channels move the student.

\begin{table}[t]
\caption{Summary of the 510-patient paired comparisons (five folds $\times$ three seeds). Unless stated otherwise, $\Delta\mathrm{BA}$ is against the batch-512 supervised baseline. $p$: paired $t$-test over the 15 fold/seed cells (patient-level bootstrap for the row so marked); $p_{\mathrm{Holm}}$: Holm-adjusted across the 16 JDCNet configurations. The two JDCNet rows marked ``selected'' were chosen post hoc from 16 configurations (Table~\ref{tab:jdcnet_510}).}
\label{tab:main_decision_summary}
\centering
\scriptsize
\setlength{\tabcolsep}{2.5pt}
\renewcommand{\arraystretch}{0.95}
\begin{tabularx}{\textwidth}{@{}>{\raggedright\arraybackslash}Xrlccc@{}}
\toprule
Comparison & $\Delta\mathrm{BA}$ & 95\% CI & $+/0/-$ & $p$ & $p_{\mathrm{Holm}}$ \\
\midrule
\multicolumn{6}{@{}l}{\emph{Does the CT view carry class signal? (CT-only reference)}} \\
\quad CT mid-slice & $+0.045$ & [$+0.019$, $+0.070$] & 11/0/4 & $0.005$ & -- \\
\quad CT 3-slice & $+0.051$ & [$+0.025$, $+0.079$] & 12/0/3 & $0.003$ & -- \\
\multicolumn{6}{@{}l}{\emph{Does JDCNet move the X-ray student? (original sweep)}} \\
\quad 3-slice, soft-KL, $\tau{=}0.70$, $\lambda{=}1.0$ (selected) & $+0.0345$ & [$+0.0112$, $+0.0571$] & 10/0/5 & $0.014$ & $0.22$ \\
\quad Mid, hard, $\tau{=}0.80$, $\lambda{=}1.5$ (selected) & $+0.0329$ & [$+0.0074$, $+0.0584$] & 10/1/4 & $0.028$ & $0.42$ \\
\quad Average of all 16 configurations (unselected) & $+0.020$ & [$+0.003$, $+0.039$] & 13/0/2 & $0.060$ & -- \\
\quad Average of the 8 initial configurations (unselected) & $+0.016$ & [$-0.001$, $+0.035$] & 8/0/7 & $0.11$ & -- \\
\multicolumn{6}{@{}l}{\emph{Pre-registered supplementary experiment (batch 128, 16 supervised restarts per cell)}} \\
\quad Supervised batch 128 ($r{=}0$) minus supervised batch 512 & $-0.005$ & [$-0.032$, $+0.022$] & 8/0/7 & $0.75$ & -- \\
\quad JDCNet best of 16 minus supervised best of 16, selected on validation (primary) & $+0.037$ & [$+0.016$, $+0.058$] & 11/1/3 & $0.005$ & -- \\
\quad Same comparison, patient-level bootstrap & $+0.037$ & [$+0.007$, $+0.068$] & -- & $0.017$ & -- \\
\quad Average of all 16 configurations minus mean of 16 restarts & $+0.019$ & [$+0.006$, $+0.031$] & 12/0/3 & $0.016$ & -- \\
\multicolumn{6}{@{}l}{\emph{Do alternative channels move the X-ray student?}} \\
\quad Gated logit KD, mid-slice & $-0.022$ & [$-0.055$, $+0.010$] & 5/0/10 & $0.23$ & -- \\
\quad Gated logit KD, DRR & $-0.064$ & [$-0.095$, $-0.034$] & 3/0/12 & $0.002$ & -- \\
\quad Plain logit KD, 3-slice & $-0.036$ & [$-0.062$, $-0.013$] & 4/0/11 & $0.016$ & -- \\
\quad Contrastive alignment, best of 4 & $+0.008$ & [$-0.021$, $+0.038$] & 7/1/7 & $0.61$ & -- \\
\bottomrule
\end{tabularx}
\end{table}

\section{Experiments}
\label{sec:experiments}

\subsection{Datasets and Cohort Construction}
\label{sec:datasets}

\textit{Ethics and informed consent.} This secondary analysis used publicly released, de-identified human imaging data. The BIMCV-COVID19 releases~\cite{delaiglesiavaya2020bimcv} report institutional ethics approval and anonymization by the data custodians. No new data were collected, no patients were contacted, and no re-identification was attempted (see Declarations).

\textit{Primary cohort.} The primary cohort contains every patient in the BIMCV-COVID19+ release (label COVID-19) and the BIMCV-COVID19$-$ release (label non-COVID) for whom at least one chest CT volume and at least one chest radiograph could be retrieved, yielding 510 patients (113 COVID-19, 397 non-COVID). Each patient contributes one radiograph and one CT volume (the largest CT file, which is typically the full-volume acquisition). All selected radiographs are anteroposterior projections. No acquisition-date window was imposed between the CT and the radiograph, so the pair is same-patient but not necessarily same-visit, and the positive and negative classes come from two different BIMCV releases. Patients were assigned to five folds by seeded random permutation within each class. For fold $k$, fold $k$ is the test partition, fold $(k{+}1)\bmod 5$ the validation partition used for checkpoint selection, and the remaining three folds the training partition (about 306 training, 102 validation, and 102 test patients). Appendix Section~\ref{sec:app_cohort_construction} gives the full rules.

Table~\ref{tab:dataset_protocol} separates the data sources used in the paper. Only the 510-patient cohort supports the main comparisons; the other sources serve feasibility, capacity-control, rerun, or stress-test roles.

\begin{table}[htbp]
\caption{Data sources and their roles. Only the 510-patient paired BIMCV cohort supports the main comparisons.}
\label{tab:dataset_protocol}
\centering
\footnotesize
\setlength{\tabcolsep}{4pt}
\renewcommand{\arraystretch}{0.95}
\begin{tabularx}{\textwidth}{|>{\raggedright\arraybackslash}p{0.20\textwidth}|>{\raggedright\arraybackslash}p{0.17\textwidth}|>{\raggedright\arraybackslash}p{0.15\textwidth}|>{\raggedright\arraybackslash}p{0.20\textwidth}|>{\raggedright\arraybackslash}X|}
\hline
Cohort / source & Size & Pairing & Role in paper & Main caveat \\ \hline
Primary BIMCV paired cohort & 510 patients ($113^+ / 397^-$) & Same-patient CT--X-ray & JDCNet sweep, CT-only references, logit-KD and contrastive comparators, batch-matched supervised restarts & Single public source; classes from two releases; binary task. \\ \hline
Balanced BIMCV subsets & 226--228 patients (1:1); 216 with a rendered CT teacher view & Same-patient CT--X-ray & CT-view comparison, BiomedCLIP control, earlier reruns (appendix) & Smaller, class-balanced; different split. \\ \hline
BIMCV rerun split & 466 patients with a rendered CT teacher view & Same-patient CT--X-ray & Earlier gate-temperature rerun (appendix) & Paired comparison on the validation partition only. \\ \hline
Historical Cohen paired subset & 19 patients / 26 X-rays & Same-patient CT--X-ray & Early feasibility and instability diagnosis & Too small for ranking. \\ \hline
External MIDRC radiographs & 126 images ($63^+/63^-$) & X-ray only & Frozen-student stress test of the 510-patient models & Different source, detector mix, labels, and population. \\ \hline
\end{tabularx}
\end{table}

Fig.~\ref{fig:paired_data_schematic} summarizes, for one patient, the two modalities and the four CT teacher views derived from the CT volume. The figure is a schematic and contains no patient images.

\begin{figure}[!t]
\centering
\resizebox{0.98\textwidth}{!}{%
\begin{tikzpicture}[
  box/.style={draw, rounded corners, align=center, font=\small, inner sep=4pt},
  ctbox/.style={box, fill=gray!12, draw=gray!60},
  xrbox/.style={box, fill=blue!8, draw=blue!55},
  ptbox/.style={box, fill=orange!8, draw=orange!60},
  lab/.style={align=left, anchor=west, font=\small, text width=44mm,
    execute at begin node={\hyphenpenalty=10000\exhyphenpenalty=10000}},
  head/.style={font=\bfseries\small, anchor=west},
  arr/.style={-{Latex[length=2mm]}, thick},
  sq/.style={draw=gray!70, line width=0.5pt},
  axial/.pic={
    \fill[gray!45] (0,0) ellipse (0.40 and 0.27);
    \fill[white] (-0.17,0.02) ellipse (0.13 and 0.17);
    \fill[white] (0.17,0.02) ellipse (0.13 and 0.17);
    \fill[gray!70] (0,-0.17) circle (0.05);
  },
  thorax/.pic={
    \fill[#1!35] (-0.42,-0.50) -- (-0.46,0.20) .. controls (-0.42,0.45) and (-0.20,0.52) .. (0,0.52)
      .. controls (0.20,0.52) and (0.42,0.45) .. (0.46,0.20) -- (0.42,-0.50) -- cycle;
    \fill[white] (-0.20,0.0) ellipse (0.15 and 0.36);
    \fill[white] (0.20,0.0) ellipse (0.15 and 0.36);
    \draw[#1!70, line width=0.6pt] (0,0.40) -- (0,-0.45);
  },
]
\node[ptbox, text width=33mm] (pt) at (-0.45,0.6) {\textbf{Patient $i$}\\[2pt]
  113 COVID-19\\\mbox{(BIMCV-COVID19+)}\\[1pt]397 non-COVID\\\mbox{(BIMCV-COVID19$-$)}};
\node[font=\footnotesize, align=center] at (-0.45,-1.3) {Same patient; not\\necessarily the same visit};

\foreach \k in {0,...,6}{
  \pgfmathsetmacro{\y}{2.05+0.2*\k}
  \ifnum\k=3
    \filldraw[draw=orange!80!black, fill=orange!55] (2.5,\y) -- (4.2,\y) -- (4.8,\y+0.55) -- (3.1,\y+0.55) -- cycle;
  \else\ifnum\k=2
    \filldraw[draw=orange!70!black, fill=orange!25] (2.5,\y) -- (4.2,\y) -- (4.8,\y+0.55) -- (3.1,\y+0.55) -- cycle;
  \else\ifnum\k=4
    \filldraw[draw=orange!70!black, fill=orange!25] (2.5,\y) -- (4.2,\y) -- (4.8,\y+0.55) -- (3.1,\y+0.55) -- cycle;
  \else
    \filldraw[draw=gray!70, fill=gray!12] (2.5,\y) -- (4.2,\y) -- (4.8,\y+0.55) -- (3.1,\y+0.55) -- cycle;
  \fi\fi\fi
}
\draw[arr, gray!70] (2.2,2.05) -- (2.2,3.55) node[left, font=\small, text=black] {$z$};
\node[font=\small] at (3.65,1.6) {Chest CT volume};

\draw[blue!55, fill=blue!5, rounded corners=2pt] (2.95,-2.55) rectangle (4.35,-0.95);
\pic at (3.65,-1.75) {thorax=blue};
\node[font=\small] at (3.65,-2.95) {Chest radiograph (AP)};

\draw[arr] (pt.north east) .. controls (1.7,2.6) .. (2.45,2.75);
\draw[arr] (pt.south east) .. controls (1.7,-1.6) .. (2.9,-1.75);

\node[ctbox, text width=21mm] (win) at (6.35,2.95) {Window to\\$[-1000,400]$\\HU; resize to\\$224{\times}224$};
\node[xrbox, text width=21mm] (rsz) at (6.35,-1.75) {Resize to\\$224{\times}224$;\\scale to $[0,1]$};
\draw[arr] (4.75,2.95) -- (win.west);
\draw[arr] (4.4,-1.75) -- (rsz.west);

\node[head] at (8.1,5.35) {CT teacher views (training only)};
\foreach \d in {0.14, 0.07, 0}{
  \filldraw[sq, fill=white] (8.3+\d,4.0+\d) rectangle (9.3+\d,5.0+\d);
  \pic at (8.8+\d,4.5+\d) {axial};
}
\node[lab] at (9.65,4.55) {\textbf{Mid-slice}: central axial slice, copied to three channels};
\foreach \d/\c in {0.14/25, 0.07/55, 0/25}{
  \filldraw[sq, fill=orange!\c] (8.3+\d,2.6+\d) rectangle (9.3+\d,3.6+\d);
}
\pic at (8.8,3.1) {axial};
\node[lab] at (9.65,3.15) {\textbf{3-slice}: central slice and slices about 5~mm above and below};
\filldraw[sq, fill=gray!20] (8.3,1.2) rectangle (9.3,2.2);
\node[font=\small] at (8.8,1.7) {$\tfrac{1}{N}\!\sum_{z}$};
\node[lab] at (9.65,1.7) {\textbf{Mean projection}: volume averaged along the axial axis};
\filldraw[sq, fill=gray!8] (8.3,-0.2) rectangle (9.3,0.8);
\pic at (8.8,0.3) {thorax=gray};
\node[lab] at (9.65,0.3) {\textbf{DRR}: rendered radiograph-like projection};
\foreach \yy in {4.5,3.1,1.7,0.3}{ \draw[arr] (win.east) -- (8.2,\yy); }

\node[head] at (8.1,-0.85) {X-ray input (training and deployment)};
\filldraw[draw=blue!55, fill=blue!5] (8.3,-2.25) rectangle (9.3,-1.25);
\pic at (8.8,-1.75) {thorax=blue};
\node[lab] at (9.65,-1.75) {\textbf{X-ray}: student input in training; the only input at deployment};
\draw[arr] (rsz.east) -- (8.2,-1.75);
\end{tikzpicture}%
}
\caption{Schematic of the paired data for one patient; no patient images are shown. Each patient of the 510-patient cohort contributes one chest CT volume and one anteroposterior (AP) chest radiograph from the same patient, not necessarily acquired at the same visit, and the COVID-19 and non-COVID patients come from two different BIMCV releases. The CT volume is windowed in Hounsfield units (HU) and rendered into four teacher views that are used only during training: JDCNet uses the mid-slice and 3-slice teachers, and all four views are used for the CT-only references and the logit-KD comparators. The radiograph is the student input during training and the only input at deployment.}
\label{fig:paired_data_schematic}
\end{figure}

\subsection{Experimental Protocol and Implementation Details}
\label{sec:implementation_details}

\textbf{Models and inputs.} Teacher and student are ImageNet-pretrained ResNet-18 classifiers with a two-class head at $224{\times}224$ input. Images are resized to $224{\times}224$ and scaled to $[0,1]$ without mean--standard deviation normalization. Training images receive a random horizontal flip ($p{=}0.5$) and a mild affine perturbation (rotation $\pm5^\circ$, translation $\pm2\%$, scale $[0.95,1.05]$). CT volumes are windowed to $[-1000,400]$ Hounsfield units (HU). The mid-slice view is the central axial slice replicated to three channels; the 3-slice view stacks the central slice and the slices about 5~mm above and below it as three channels; the mean-projection view averages the windowed volume along the axial axis; and the DRR view is a rendered radiograph-like projection.

\textbf{Optimization and selection.} All supervised, CT-teacher, JDCNet, and logit-KD runs use AdamW (learning rate $3\times10^{-4}$, weight decay $10^{-4}$) for 50 epochs without a learning-rate schedule (the BiomedCLIP control used a learning rate of $10^{-5}$, and contrastive pretraining ran for 100 epochs), mixed-precision training with 16-bit floating point (FP16), inverse-frequency class-weighted cross-entropy, and a class-weighted sampler. The checkpoint with the highest validation BA is evaluated once on the test partition. Batch size is 128 for JDCNet, the contrastive comparator, and the supplementary supervised restarts, and 512 for the original supervised baseline, CT-only references, and logit-KD comparators (Section~\ref{sec:statistical_protocol}).

\textbf{CT teachers are trained within each fold.} For every fold and seed, a separate CT teacher for each view is trained only on that fold's training partition, selected on its validation partition, frozen, and used to supervise the student of the same fold and seed. Teachers never see test-partition patients, so no test labels or images leak through the teacher. JDCNet reuses the teachers trained for the CT-only reference rows (15 per view).

\textbf{Hardware and cost.} The 510-patient runs used NVIDIA RTX 3090 GPUs; the initial 120-run JDCNet stage required about 32 minutes on two GPUs. Per optimizer step, JDCNet adds one forward pass of the frozen ResNet-18 teacher to the student's forward and backward passes, an analytic increase of roughly one third in training computation at equal batch size; wall-clock training overhead was not recorded. At inference the teacher, gate, and auxiliary loss are discarded, so the per-image cost is that of the ResNet-18 student (Table~\ref{tab:efficiency}).

\subsection{Baselines and Comparators}

The primary baseline is the supervised X-ray ResNet-18 trained on the same 510-patient folds. The CT-only reference rows use the same architecture on CT views; they indicate whether the privileged modality carries class signal but are not deployable X-ray models. Transfer comparators on the 510-patient splits are plain logit KD, gated logit KD, and contrastive alignment (Section~\ref{sec:comparators}). Late fusion was evaluated only in the early feasibility stage because it requires CT at inference.

\subsection{JDCNet Sweep on the 510-Patient Cohort}
\label{sec:jdcnet_results_main}

Table~\ref{tab:jdcnet_510} lists all 16 JDCNet configurations in order of mean $\Delta\mathrm{BA}$, with the exploration stage, the fold/seed win counts, directly computed $p$-values, and the difference from the mean of 16 batch-matched supervised restarts of the same cell (Section~\ref{sec:batch_matched_reruns}).

\begin{table}[htbp]
\caption{All 16 JDCNet configurations on the 510-patient BIMCV cohort (ResNet-18 student; five folds $\times$ seeds 42--44). Stage: I, initial grid; A, extension A ($\lambda{=}1.5$); B, extension B (soft-KL), in the order they were run. $+/0/-$: fold/seed cells with positive, zero, and negative $\Delta\mathrm{BA}$ against the batch-512 supervised baseline. CI: recorded percentile bootstrap. $p$: paired $t$-test over the 15 cells; $p_{\mathrm{H}}$: Holm-adjusted over the 16 configurations. $\Delta_{128}$: mean difference from the average of the 16 batch-128 supervised restarts of the same cell. Rule: meets the planned decision rule (mean $\geq+0.03$ and lower CI bound $>0$).}
\label{tab:jdcnet_510}
\centering
\scriptsize
\setlength{\tabcolsep}{1.6pt}
\renewcommand{\arraystretch}{0.9}
\begin{tabular}{@{}lcccccrlcccc@{}}
\toprule
Teacher & $\tau$ & $\lambda$ & Target & Stage & $+/0/-$ & $\Delta\mathrm{BA}$ & 95\% CI & $p$ & $p_{\mathrm{H}}$ & $\Delta_{128}$ & Rule \\
\midrule
3-slice & 0.70 & 1.0 & soft-KL & B & 10/0/5 & $+0.0345$ & [$+0.0112$, $+0.0571$] & 0.014 & 0.22 & $+0.033$ & Yes \\
Mid & 0.80 & 1.5 & hard & A & 10/1/4 & $+0.0329$ & [$+0.0074$, $+0.0584$] & 0.028 & 0.42 & $+0.032$ & Yes \\
Mid & 0.70 & 1.0 & hard & I & 10/0/5 & $+0.0298$ & [$-0.0002$, $+0.0597$] & 0.079 & 1.00 & $+0.029$ & -- \\
Mid & 0.70 & 1.5 & hard & A & 9/0/6 & $+0.0296$ & [$+0.0026$, $+0.0577$] & 0.061 & 0.86 & $+0.029$ & -- \\
Mid & 0.70 & 1.0 & soft-KL & B & 10/0/5 & $+0.0264$ & [$+0.0017$, $+0.0547$] & 0.082 & 1.00 & $+0.025$ & -- \\
3-slice & 0.80 & 1.0 & soft-KL & B & 10/0/5 & $+0.0258$ & [$-0.0029$, $+0.0546$] & 0.11 & 1.00 & $+0.025$ & -- \\
3-slice & 0.80 & 1.0 & hard & I & 10/0/5 & $+0.0247$ & [$+0.0012$, $+0.0504$] & 0.082 & 1.00 & $+0.024$ & -- \\
3-slice & 0.70 & 0.5 & hard & I & 10/0/5 & $+0.0239$ & [$-0.0032$, $+0.0501$] & 0.11 & 1.00 & $+0.023$ & -- \\
Mid & 0.80 & 1.0 & soft-KL & B & 9/0/6 & $+0.0231$ & [$-0.0083$, $+0.0544$] & 0.19 & 1.00 & $+0.022$ & -- \\
3-slice & 0.80 & 1.5 & hard & A & 9/0/6 & $+0.0181$ & [$-0.0105$, $+0.0464$] & 0.25 & 1.00 & $+0.017$ & -- \\
3-slice & 0.80 & 0.5 & hard & I & 9/0/6 & $+0.0151$ & [$-0.0185$, $+0.0479$] & 0.40 & 1.00 & $+0.014$ & -- \\
Mid & 0.80 & 0.5 & hard & I & 10/0/5 & $+0.0144$ & [$-0.0063$, $+0.0358$] & 0.22 & 1.00 & $+0.013$ & -- \\
Mid & 0.80 & 1.0 & hard & I & 10/0/5 & $+0.0127$ & [$-0.0177$, $+0.0423$] & 0.44 & 1.00 & $+0.012$ & -- \\
3-slice & 0.70 & 1.0 & hard & I & 7/0/8 & $+0.0066$ & [$-0.0182$, $+0.0330$] & 0.63 & 1.00 & $+0.005$ & -- \\
Mid & 0.70 & 0.5 & hard & I & 7/0/8 & $+0.0005$ & [$-0.0247$, $+0.0264$] & 0.97 & 1.00 & $-0.001$ & -- \\
3-slice & 0.70 & 1.5 & hard & A & 9/0/6 & $-0.0031$ & [$-0.0315$, $+0.0253$] & 0.84 & 1.00 & $-0.004$ & -- \\
\bottomrule
\end{tabular}
\end{table}

\textbf{Main result.} Fifteen of 16 configurations have a positive mean $\Delta\mathrm{BA}$ against the batch-512 baseline. The per-cell average over all 16 configurations is $+0.020$ (95\% CI $[+0.003,+0.039]$; paired $t$-test $p=0.060$), and over the eight initial configurations $+0.016$ ($[-0.001,+0.035]$; $p=0.11$). Two configurations met the planned decision rule: 3-slice soft-KL ($\tau{=}0.70$, $\lambda{=}1.0$; $+0.0345$, CI $[+0.0112,+0.0571]$) and mid-slice hard ($\tau{=}0.80$, $\lambda{=}1.5$; $+0.0329$, CI $[+0.0074,+0.0584]$), both from the extension stages.

\textbf{Selection-adjusted interpretation.} The paired $t$-test gives $p=0.014$ and $p=0.028$ for these two configurations. After Holm adjustment over 16 configurations they become $0.22$ and $0.42$, and Benjamini--Hochberg adjustment gives $0.22$ for both. No individual configuration is therefore significant after multiplicity adjustment, and the estimates of the two selected configurations are likely optimistic. Descriptively, and without attached uncertainty, the two selected estimates correspond to $0.68$ and $0.73$ of the CT-only reference differences of the same teacher views, and the 16-configuration average to $0.41$ of the mean reference difference. Because numerators and denominators are uncertain and the numerators were selected, these ratios are not estimates of the transferable fraction of CT signal. Whether JDCNet as a method, rather than a particular configuration, improves BA is addressed by the pre-registered nested comparison of Section~\ref{sec:batch_matched_reruns}, which applies the same best-of-16 selection to the supervised baseline.

\textbf{Sensitivity to $\tau_{\mathrm{gate}}$ and $\lambda$, and risk of overfitting.} Changes across $\tau$ and $\lambda$ can look large when individual configurations are compared, but they are small relative to the sampling noise. Holding view, target, and $\lambda$ fixed, moving $\tau$ from $0.70$ to $0.80$ changes $\Delta\mathrm{BA}$ by at most $0.021$. The standard deviation of the 16 configuration means ($0.011$) is smaller than their typical standard error ($0.014$) and than the approximate standard error of a difference between two configurations ($0.016$), so the observed spread is compatible with a common underlying effect. For comparison, 16 supervised restarts that differed only in the random seed varied with a within-cell standard deviation of $0.041$ BA (Section~\ref{sec:batch_matched_reruns}), so run-to-run variability, rather than $\tau$ or $\lambda$, dominates the spread of individual runs. The hard-target $\lambda$ sweeps are non-monotonic and teacher-dependent (mid-slice, $\tau{=}0.80$: $+0.014\rightarrow+0.013\rightarrow+0.033$; 3-slice, $\tau{=}0.70$: $+0.024\rightarrow+0.007\rightarrow-0.003$ for $\lambda=0.5,1.0,1.5$). Model-level overfitting to the test partition is limited because checkpoints were selected on a separate validation fold and teachers were trained within fold. The substantive risk is selection overfitting: $\tau$, $\lambda$, and the target type were chosen by looking at test-partition results on the same 510 patients. The nested comparison in Section~\ref{sec:batch_matched_reruns} removes the use of test data for this choice; an independent cohort is still needed to remove the dependence on these patients.

\textbf{Hard versus soft-KL target.} At matched teacher view, $\tau$, and $\lambda{=}1.0$, the soft-KL target minus the hard target gives $+0.028$ (3-slice, $\tau{=}0.70$), $+0.001$ (3-slice, $0.80$), $-0.003$ (mid, $0.70$), and $+0.010$ (mid, $0.80$), an average of $+0.009$. Retaining the teacher's confidence magnitude may help slightly, but the differences are within the sampling uncertainty, consistent with the small amount of information the hard target discards (Section~\ref{sec:theoretical_rationale}).

\textbf{Absolute operating points.} Table~\ref{tab:app_absolute_metrics} reports absolute test-partition metrics. The supervised baseline reaches BA $0.625$ and AUC $0.674$ at batch size 512, and $0.620$ and $0.668$ at batch size 128. The two selected configurations reach BA $0.659$ and $0.658$ and AUC $0.713$ and $0.692$. Their gains come through specificity ($0.711$ and $0.674$ versus $0.620$) with little change in sensitivity ($0.607$ and $0.642$ versus $0.629$); across all 16 configurations the mean change in specificity was positive ($+0.013$ to $+0.091$) and the mean change in sensitivity was negative in 12 (range over all 16: $-0.077$ to $+0.016$; Appendix Table~\ref{tab:app_initial_grid_tests}). The CT-only references have the same profile (specificity $0.81$--$0.82$, sensitivity $0.53$), so JDCNet moves the student's operating point toward that of the teacher. A higher BA at this retrospective operating point does not by itself indicate better clinical triage.

\begin{table}[htbp]
\caption{Absolute test-partition performance on the 510-patient cohort, mean (SD) over 15 fold/seed cells. Macro-F1 is the unweighted mean of the per-class F1 scores; sensitivity refers to the COVID-19 class.}
\label{tab:app_absolute_metrics}
\centering
\scriptsize
\setlength{\tabcolsep}{2.2pt}
\renewcommand{\arraystretch}{0.9}
\begin{tabularx}{\textwidth}{@{}>{\raggedright\arraybackslash}Xccccc@{}}
\toprule
Method & BA & AUC & Macro-F1 & Sensitivity & Specificity \\
\midrule
Supervised X-ray, batch 512 & $0.625\,(0.050)$ & $0.674\,(0.049)$ & $0.568$ & $0.629\,(0.139)$ & $0.620\,(0.128)$ \\
Supervised X-ray, batch 128 ($r{=}0$) & $0.620\,(0.066)$ & $0.668\,(0.053)$ & $0.558$ & $0.646\,(0.130)$ & $0.594\,(0.127)$ \\
JDCNet 3-slice soft-KL & $0.659\,(0.041)$ & $0.713\,(0.027)$ & $0.618$ & $0.607\,(0.153)$ & $0.711\,(0.125)$ \\
JDCNet mid hard & $0.658\,(0.038)$ & $0.692\,(0.045)$ & $0.607$ & $0.642\,(0.138)$ & $0.674\,(0.132)$ \\
Plain logit KD, mid & $0.614\,(0.071)$ & $0.676\,(0.072)$ & $0.565$ & $0.550\,(0.214)$ & $0.679\,(0.228)$ \\
Gated logit KD, mid & $0.603\,(0.079)$ & $0.660\,(0.068)$ & $0.528$ & $0.558\,(0.290)$ & $0.647\,(0.302)$ \\
Gated logit KD, DRR & $0.561\,(0.059)$ & $0.613\,(0.077)$ & $0.429$ & $0.689\,(0.287)$ & $0.433\,(0.324)$ \\
CT-only reference, mid & $0.670\,(0.051)$ & $0.747\,(0.050)$ & $0.653$ & $0.531\,(0.155)$ & $0.809\,(0.097)$ \\
CT-only reference, 3-slice & $0.676\,(0.045)$ & $0.754\,(0.032)$ & $0.662$ & $0.534\,(0.136)$ & $0.818\,(0.096)$ \\
Contrastive alignment, best & $0.633\,(0.047)$ & $0.661\,(0.052)$ & $0.577$ & $0.642\,(0.090)$ & $0.623\,(0.092)$ \\
\bottomrule
\end{tabularx}
\end{table}

\textbf{Teacher reliability and gate coverage.} The CT views were re-rendered from the CT volumes with the original rendering code, and the re-run teachers reproduced the recorded teacher test BA exactly, so the teacher's behavior on each partition can be examined (Table~\ref{tab:teacher_reliability}). Two observations bear on the mechanism. First, the gate acts on the training partition, on which each teacher was itself trained. There the mid-slice and 3-slice teachers are 94--96\% accurate and nearly calibrated (ECE $0.02$--$0.03$); the gate retains 84\% (mid-slice, $\tau{=}0.80$) and 95\% (3-slice, $\tau{=}0.70$) of training samples; and about 97\% of retained samples receive an argmax target equal to the ground-truth label. The hard auxiliary target therefore mostly repeats the true label on samples that the CT teacher classifies confidently and contradicts it on about 3\% of retained samples. Its main effect is a CT-informed reweighting of training samples rather than new label information; the soft-KL target additionally carries the teacher's confidence. Second, on held-out patients the teachers are over-confident (validation and test ECE $0.17$--$0.19$; validation-fitted temperature $1.7$), so held-out calibration does not describe the regime in which the gate operates. Calibrating with the validation-fitted temperature would move the effective raw-confidence threshold of the mid-slice configuration from $0.80$ to $0.92$ and that of the 3-slice configuration from $0.70$ to $0.81$ (Eq.~\eqref{eq:temperature_threshold}). A teacher's confidence on its own training patients is optimistic; cross-fitted teachers, whose predictions on a training sample come from a model that did not see it, are a natural refinement.

\begin{table}[htbp]
\caption{CT-teacher reliability by partition (mean over the 15 fold/seed teachers of each view). Cov.: fraction of samples with teacher confidence above $\tau$; Acc.\ ret./rej.: teacher accuracy on samples with confidence above/below $\tau{=}0.80$; ECE: expected calibration error (15 bins); $T_{\mathrm{val}}$: temperature fitted on the validation partition by negative log-likelihood. The gate acts on the training partition.}
\label{tab:teacher_reliability}
\centering
\scriptsize
\setlength{\tabcolsep}{2.4pt}
\renewcommand{\arraystretch}{0.9}
\begin{tabular}{@{}llcccccc@{}}
\toprule
View & Partition & Acc. & ECE & Cov.\ $\tau{=}0.70$ & Cov.\ $\tau{=}0.80$ & Acc.\ ret./rej. & $T_{\mathrm{val}}$ \\
\midrule
Mid-slice & train & 0.943 & 0.032 & 0.89 & 0.84 & 0.965 / 0.700 & 1.72 \\
 & val & 0.787 & 0.169 & 0.83 & 0.76 & 0.828 / 0.689 & 1.72 \\
 & test & 0.747 & 0.189 & 0.82 & 0.75 & 0.772 / 0.568 & 1.72 \\
3-slice & train & 0.962 & 0.021 & 0.95 & 0.92 & 0.981 / 0.727 & 1.71 \\
 & val & 0.789 & 0.174 & 0.89 & 0.82 & 0.813 / 0.674 & 1.71 \\
 & test & 0.755 & 0.194 & 0.90 & 0.82 & 0.792 / 0.551 & 1.71 \\
Mean proj. & train & 0.881 & 0.057 & 0.84 & 0.74 & 0.920 / 0.688 & 2.52 \\
 & val & 0.764 & 0.176 & 0.79 & 0.68 & 0.807 / 0.687 & 2.52 \\
 & test & 0.723 & 0.190 & 0.79 & 0.69 & 0.765 / 0.523 & 2.52 \\
DRR & train & 0.865 & 0.060 & 0.84 & 0.77 & 0.937 / 0.590 & 1.15 \\
 & val & 0.746 & 0.194 & 0.80 & 0.71 & 0.732 / 0.681 & 1.15 \\
 & test & 0.699 & 0.202 & 0.79 & 0.71 & 0.768 / 0.524 & 1.15 \\
\bottomrule
\end{tabular}
\end{table}

\textbf{Gate behavior on individual cases.} Table~\ref{tab:gate_case_examples} lists four anonymized training cases with the output of a mid-slice CT teacher and the gate at $\tau_{\mathrm{gate}}=0.80$. Acceptance is not correctness: Case~B is a confidently wrong prediction that the gate admits, and Case~C is a correct prediction that the gate rejects. Such admitted errors cannot be removed by temperature scaling (Eq.~\eqref{eq:temperature_threshold}).

\begin{table}[t]
\caption{Anonymized case-level outputs of a mid-slice CT teacher and the confidence gate ($\tau_{\mathrm{gate}}=0.80$; fold~0, seed~42). Case labels do not encode patient identifiers, and no pixels, filenames, or identifiers are reproduced. The source collection is the BIMCV-COVID19 portal, \url{https://bimcv.cipf.es/bimcv-projects/bimcv-covid19}. The four examples are illustrative, not a performance estimate.}
\label{tab:gate_case_examples}
\centering
\scriptsize
\setlength{\tabcolsep}{3pt}
\renewcommand{\arraystretch}{0.95}
\begin{tabular}{@{}lcccc@{}}
\toprule
Case & Teacher confidence & Teacher prediction & Ground truth & Gate outcome \\
\midrule
Case A & 0.9992 & COVID-19 & COVID-19 & Accepted \\
Case B & 0.9997 & Non-COVID & COVID-19 & Accepted \\
Case C & 0.7619 & COVID-19 & COVID-19 & Rejected \\
Case D & 0.7707 & Non-COVID & COVID-19 & Rejected \\
\bottomrule
\end{tabular}
\end{table}

\subsection{Comparator Results on the Same Cohort}
\label{sec:comparator_audit_main}

Under the same 510-patient splits, no comparator transfer mechanism showed a gain (Table~\ref{tab:main_decision_summary}; Appendix Table~\ref{tab:app_comparator_summary}). Gated logit KD was negative or neutral across the four CT teacher views: significantly worse than the baseline for the DRR view ($-0.064$, CI $[-0.095,-0.034]$) and below it with CIs including zero for the mid-slice ($-0.022$, $[-0.055,+0.010]$), 3-slice ($-0.026$, $[-0.078,+0.028]$), and mean-projection views ($-0.014$, $[-0.056,+0.030]$). Plain logit KD was also negative or neutral for every view, significantly so for the 3-slice view ($-0.036$, $[-0.062,-0.013]$; mid-slice $-0.010$, mean projection $-0.016$, DRR $-0.001$). Contrastive alignment remained near zero for all four configurations (best $+0.008$, CI crossing zero; Appendix Table~\ref{tab:app_contrastive_510}). On a 226-patient balanced subset, a BiomedCLIP ViT-B/16 student was not distinguishable from the ResNet-18 supervised baseline trained on the same subset ($+0.002$, CI $[-0.048,+0.050]$), so a single configuration of a larger backbone showed no detectable gain there.

The comparator search was much smaller than the JDCNet search: each logit-KD comparator was run with one setting ($T{=}4$, $\alpha{=}0.6$) per teacher view on the 510-patient cohort, contrastive alignment with four settings, and BiomedCLIP with one setting (learning rate $10^{-5}$, unfrozen encoder). An earlier scan of gated logit KD over $T\in\{2,4,8\}$ and gate thresholds $\{0.50,0.55,0.60\}$ on a class-balanced BIMCV subset, with a different split and training configuration, produced a best configuration of $+0.034$ with a CI crossing zero (Appendix Section~\ref{sec:app_calibration_scan}). Because the comparators were tuned less and the logit-KD comparators were trained only at batch size 512, the comparator results show that these channels did not help under the tested settings; they do not establish that JDCNet's channel is superior.

\subsection{CT-Only Reference}
\label{sec:teacher_upperbound_main}

A transfer claim presupposes that CT carries class signal beyond X-ray for the same patients. On the 510-patient cohort, the mid-slice and 3-slice CT-only references exceed the supervised X-ray baseline ($+0.045$, CI $[+0.019,+0.070]$, 11 of 15 cells positive; $+0.051$, CI $[+0.025,+0.079]$, 12 of 15), whereas the mean-projection and DRR references do not ($+0.031$, CI $[-0.002,+0.066]$; $+0.010$, CI $[-0.036,+0.055]$). JDCNet was therefore run with the mid-slice and 3-slice teachers only. The CT-only reference is not a mathematical upper bound on the student: an X-ray student could in principle exceed it, and the reference itself is a selected, uncertain estimate. The DRR-guided gated logit KD result is consistent with geometric similarity to a radiograph not being sufficient to make soft CT logits a useful target. Appendix Table~\ref{tab:app_ct_variants} reports the view comparison on the 226-patient subset.

\subsection{Pre-Registered Supplementary Experiment: Batch-Matched Restarts and Nested Selection}
\label{sec:batch_matched_reruns}

All 240 supervised restarts of the supplementary experiment (Section~\ref{sec:statistical_protocol}) completed, and the analyses below follow the plan written before the runs (Appendix Section~\ref{sec:app_supplementary}).

\textbf{Training budget.} The batch-matched baseline (first restart, batch 128) reached a mean test BA of $0.620$, against $0.625$ at batch 512 ($-0.005$, CI $[-0.032,+0.022]$, $p=0.75$), and the mean over all 16 restarts was $0.626$. Three times as many optimizer updates did not improve supervised training on average, so the batch-size difference is unlikely to explain the JDCNet results. Against the first restart, the pre-specified batch-matched baseline, all 16 configurations had a positive mean difference ($+0.002$ to $+0.039$; average $+0.024$, CI $[-0.000,+0.051]$, $p=0.092$), and none remained significant after Holm adjustment at the cell or the patient level (smallest adjusted $p$: $0.56$ and $0.33$; Appendix Table~\ref{tab:app_vs_r0}). Against the mean of the 16 restarts of the same cell, the average over the 16 JDCNet configurations was $+0.019$ (CI $[+0.006,+0.031]$, 12 of 15 cells positive, $p=0.016$), and the two selected configurations were $+0.033$ ($p=0.028$) and $+0.032$ ($p=0.007$); Holm-adjusted over the 16 configurations these are $0.42$ and $0.12$.

\textbf{Run-to-run variability.} Restarts that differed only in the random seed varied with a within-cell standard deviation of $0.041$ BA. Differences of $0.02$--$0.04$ between two single runs are therefore within the range produced by the seed alone, which is why single-run comparisons and best-of-many selections need a matched control.

\textbf{Nested selection with a matched control (primary analysis).} When each method selected one of its 16 runs per cell on the validation partition, JDCNet exceeded supervised training by $+0.037$ BA (CI $[+0.016,+0.058]$; 11 cells positive, 1 tied, 3 negative; $p=0.005$). The patient-level bootstrap gave the same estimate with a wider interval ($+0.037$, CI $[+0.007,+0.068]$, $p=0.017$). Restricting JDCNet to the eight initial configurations and the supervised baseline to its first eight restarts gave $+0.029$ (CI $[+0.007,+0.049]$, $p=0.022$). Best-of-16 selection improved supervised training only slightly ($+0.013$ over the first restart, CI $[-0.011,+0.038]$), so giving the baseline the same number of attempts does not reproduce the JDCNet advantage. At the patient level, the two selected configurations exceeded the batch-matched baseline by $+0.039$ (CI $[+0.003,+0.077]$) and $+0.038$ ($[+0.001,+0.073]$).

\textbf{Reading.} The plan specified that a nested difference with a CI excluding zero would indicate an advantage not explained by best-of-16 selection or batch size. That condition is met. The evidence remains within one cohort: the configurations and restarts were evaluated on the same 510 patients that motivated JDCNet, the validation partition was used both for checkpoint selection and for configuration selection (equally for both methods), and the external test below shows no transfer. Two earlier reruns on a different host, which used validation-partition metrics and partly approximated CT views, are reported in Appendix Section~\ref{sec:app_gate_temperature}; this experiment supersedes them.

\subsection{Historical Feasibility Reference}
\label{sec:resampling_226_historical}

\begin{figure}[t]
\centering
\includegraphics[width=\columnwidth]{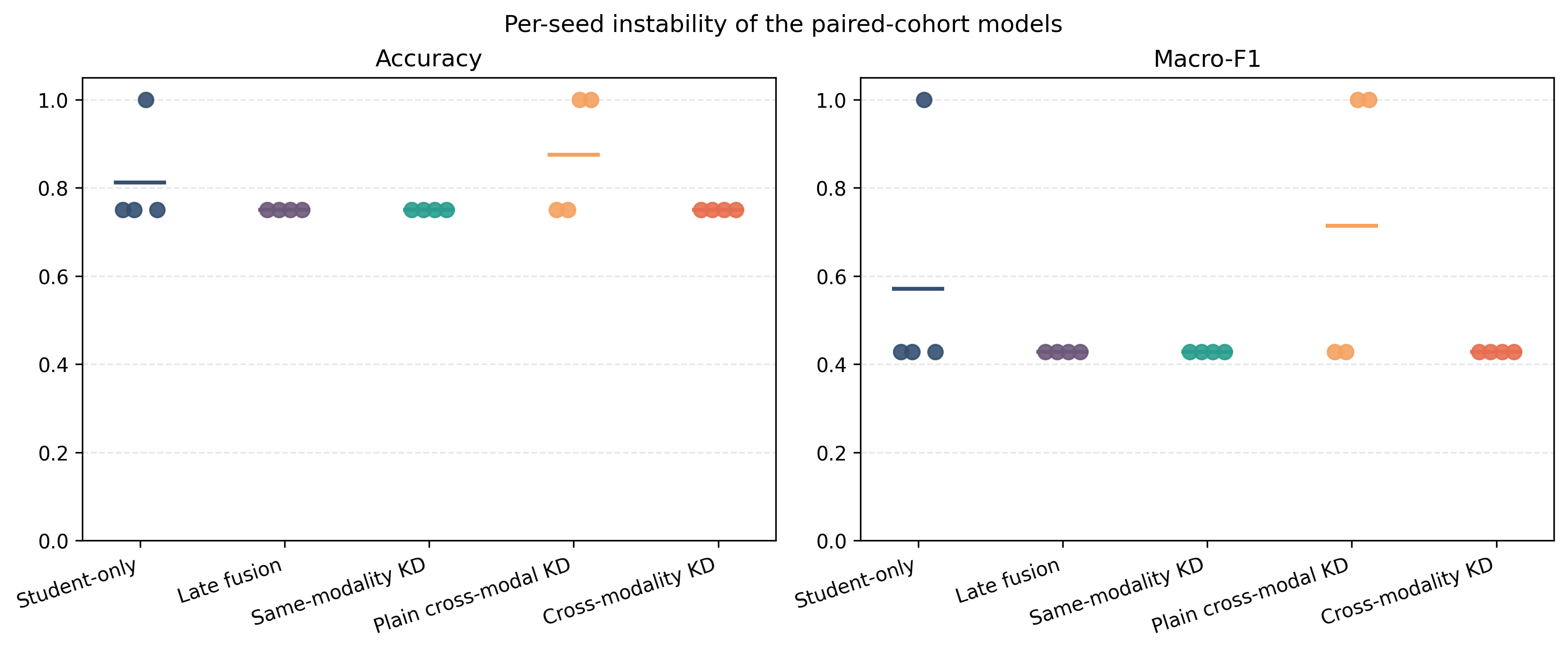}
\caption{Per-seed instability on the historical 19-patient paired cohort (left: accuracy; right: macro-F1). Each dot is one seed and the bar is the per-method mean. Plain cross-modal KD and the student-only control swing widely across seeds, so a single seed can produce a misleading ranking. This diagnostic motivates the larger cohort and the paired multi-seed protocol; it is not evidence for or against JDCNet.}
\label{fig:seed_instability}
\end{figure}

The earliest experiments used a 19-patient paired subset of the public COVID-19 image data collection~\cite{cohen2020covid}, lightweight CNNs evaluated with repeated seeds on a fixed split and later with ten patient-level Monte Carlo resamples, and several comparators (plain logit KD, attention transfer, feature hints, the module-augmented pilot, and generic KD baselines). They are retained only to explain why the larger cohort and a fixed evaluation protocol were needed. Figure~\ref{fig:seed_instability} shows that on this cohort per-seed accuracy and macro-F1 swing between chance level and perfect scores, so a single favorable seed can produce a misleading ranking. Similarly, the best configuration of an earlier gated-logit scan on a balanced BIMCV subset ($+0.034$, CI crossing zero) was not reproduced on the 510-patient cohort, where gated logit KD was negative or neutral for all four teacher views. Table~\ref{tab:stress_summary_main} summarizes these supporting layers.

\begin{table}[t]
\caption{Supporting evidence layers and what each contributes.}
\label{tab:stress_summary_main}
\centering
\footnotesize
\setlength{\tabcolsep}{3pt}
\renewcommand{\arraystretch}{0.95}
\begin{tabularx}{\columnwidth}{|>{\raggedright\arraybackslash}p{0.30\columnwidth}|>{\raggedright\arraybackslash}p{0.36\columnwidth}|>{\raggedright\arraybackslash}X|}
\hline
Evidence layer & Main observation & Role \\ \hline
19-patient paired cohort & Seed-level swings between chance and perfect scores & Feasibility only \\ \hline
Balanced-subset gated-logit scan & Best configuration positive, CI crosses zero & Instability diagnosis \\ \hline
510-patient logit-KD runs & Negative or neutral across four teacher views & Comparator evidence \\ \hline
226-patient BiomedCLIP control & No detectable difference from ResNet-18 ($+0.002$, CI $\pm0.05$) & Capacity control \\ \hline
Pre-registered restarts & No detectable batch-size effect; nested JDCNet advantage $+0.037$ & Selection and budget control \\ \hline
External MIDRC test & All students at chance & Generalization warning \\ \hline
\end{tabularx}
\end{table}

\subsection{Deployment-Time Cost}
\label{sec:deployment_efficiency}

The supervised baseline, every JDCNet configuration, and the logit-KD comparators share the same deployed model, an ImageNet-initialized ResNet-18 classifier at $224{\times}224$; the CT teacher, gate, and auxiliary loss are discarded before deployment. Table~\ref{tab:efficiency} reports parameters and MACs of this graph and of the historical module-augmented student, together with an illustrative latency and memory measurement. Because the JDCNet and supervised rows are the same computation graph, they have identical parameters, MACs, and per-image latency under any benchmark; this structural identity, rather than the particular timing values, is the conclusion supported here. It does not establish equal end-to-end deployment cost, calibration, safety, or clinical utility, which depend on the trained weights and the surrounding workflow.

\begin{table}[t]
\caption{Inference cost at $224{\times}224$ input in 32-bit floating point (FP32), batch size 1, model forward pass only. MACs count convolution and linear layers. Latency is the median (interquartile range) of 1{,}000 timed passes after 50 warm-ups on a laptop with an Intel Core i7-13700H CPU (one thread) and an NVIDIA GeForce RTX 4050 Laptop GPU (PyTorch 2.5.1); a repeated run differed by up to a factor of two on the GPU because of power management, so timings are illustrative. The supervised, JDCNet, and logit-KD students share one deployed graph and therefore one row.}
\label{tab:efficiency}
\centering
\scriptsize
\setlength{\tabcolsep}{2.4pt}
\renewcommand{\arraystretch}{0.9}
\begin{tabularx}{\columnwidth}{@{}>{\raggedright\arraybackslash}Xcccc@{}}
\toprule
Deployed model & Params / MACs & CPU latency & GPU latency & GPU peak mem. \\
\midrule
ResNet-18 X-ray classifier (supervised, JDCNet, logit KD) & 11.18\,M / 1.81\,G & 55.4 (5.5)\,ms & 2.63 (0.45)\,ms & 62\,MB \\
Module-augmented student (historical pilot) & 0.57\,M / 3.02\,G & 99.8 (11.1)\,ms & 1.93 (0.29)\,ms & 34\,MB \\
\bottomrule
\end{tabularx}
\end{table}

JDCNet is not a fusion model. At training time it costs one additional frozen-teacher forward pass per sample and requires same-patient CT for the training cohort. The module-augmented student has far fewer parameters but more MACs than ResNet-18 at $224{\times}224$ and a higher single-thread CPU latency, so parameter count alone is a poor proxy for deployment cost. Energy, end-to-end throughput, and edge-device latency were not measured. Lightweight student backbones such as MobileNetV2 and EfficientNet-B0 are natural targets for the same objective but were not evaluated in this study.

\subsection{External-Cohort Stress Test: Frozen-Student Inference}
\label{sec:external_stress}

To probe generalization beyond BIMCV, the frozen 510-patient X-ray students were evaluated without retraining or threshold adjustment on 126 class-balanced chest radiographs ($63$ COVID-19, $63$ non-COVID; one radiograph per patient) from a locked validation split of an MIDRC collection~\cite{midrc}, a source disjoint from BIMCV. For each method, all 15 fold/seed checkpoints were evaluated with per-image predictions. Table~\ref{tab:external} reports the checkpoint-averaged BA and AUC with image-level 95\% CIs (Wilson intervals for BA, which equals accuracy with balanced classes; Hanley--McNeil intervals for AUC), the AUC of the checkpoint-averaged score, and the fraction of images predicted positive.

\begin{table}[t]
\caption{External stress test of the 510-patient models on 126 MIDRC radiographs (63/63). Values are means over the 15 fold/seed checkpoints of each method; 95\% CIs reflect image-level sampling. Ens.\ AUC: AUC of the checkpoint-averaged score. Pos.\ rate: fraction of images predicted COVID-19.}
\label{tab:external}
\centering
\scriptsize
\setlength{\tabcolsep}{2.4pt}
\renewcommand{\arraystretch}{0.92}
\begin{tabularx}{\columnwidth}{@{}>{\raggedright\arraybackslash}Xcccc@{}}
\toprule
Students (checkpoints) & BA [95\% CI] & AUC [95\% CI] & Ens.\ AUC & Pos.\ rate \\
\midrule
Supervised, batch 512 (15) & 0.514 [0.427, 0.599] & 0.505 [0.404, 0.606] & 0.517 & 0.68 \\
Supervised, batch 128 (15) & 0.498 [0.412, 0.584] & 0.489 [0.387, 0.590] & 0.477 & 0.57 \\
JDCNet 3-slice soft-KL (15) & 0.492 [0.406, 0.578] & 0.498 [0.397, 0.599] & 0.485 & 0.67 \\
JDCNet mid hard (15) & 0.503 [0.417, 0.589] & 0.511 [0.409, 0.612] & 0.507 & 0.60 \\
\bottomrule
\end{tabularx}
\end{table}

Every model performs at chance: all BA and AUC intervals include $0.5$. Paired DeLong tests on the per-image, checkpoint-averaged scores show no detectable difference between JDCNet and supervised students (3-slice soft-KL minus batch-512 supervised: $\Delta\mathrm{AUC}=-0.032$, CI $[-0.087,+0.023]$, $p=0.26$; mid-slice hard: $-0.009$, CI $[-0.065,+0.047]$, $p=0.75$), and the mean cell-wise BA differences lie within $\pm0.03$ with CIs including zero. Pre-specified McNemar tests per checkpoint pair (exact test on the discordant images) gave $p<0.05$ in 2 of 15 pairs against the batch-512 students, both favoring the supervised model, and in none of 15 pairs against the batch-128 students. Because AUC is also at chance, the failure is not a recalibratable threshold offset: neither supervised nor CT-supervised students rank external positives above negatives. The data show that all tested students fail on this external set; they do not identify why. Possible explanations include the following. (a)~Acquisition differences that a single-source model may exploit as shortcuts~\cite{degrave2021shortcut,ongly2024shortcut}. In the training cohort, computed radiography (CR) made up 62\% of the positive and 45\% of the negative radiographs ($\chi^2$ test $p=0.002$), with the rest digital radiography (DX), whereas 85\% of the external radiographs are DX; detector-related appearance is thus associated with the training label but not with the external one. (b)~Differences in label definition between the external labels and the BIMCV release-based labels, which we could not independently verify. (c)~A source confound in the training data, since BIMCV positives and negatives come from two different releases. (d)~Differences in disease prevalence, severity, and patient population~\cite{ahluwalia2023subgroup}. (e)~Preprocessing differences. Single-source training is known to generalize poorly to unseen distributions even when internal validation is good~\cite{zhang2023when,sun2022performance}. The external test therefore bounds the within-cohort result rather than refuting or supporting it: the CT supervision did not confer external robustness, and robustness must be established on independent data. An earlier external evaluation of students trained on a smaller BIMCV subset gave the same result (BA $0.49$--$0.50$).

\subsection{Data Availability and Ethics}

This study is a secondary analysis of publicly released, de-identified imaging data. We follow the source-release terms, do not redistribute pixel data, and release split definitions, code, the per-run aggregate records, and the analysis scripts (see Declarations). The cell-level statistics can be recomputed from the released records; the patient-level, external, and teacher analyses additionally require per-image predictions, which are available on request because they are indexed by source-dataset identifiers. The model is not intended for clinical decision-making.

\subsection{Limitations and Future Work}
\label{sec:limitations}

\textbf{Selection and confirmatory status.} The JDCNet configurations were explored adaptively on the same 510 patients used to estimate their performance, after the cohort had been used for other comparisons. No individual configuration is significant after multiplicity adjustment. The pre-registered nested comparison removes the use of test data for configuration choice and gives the supervised baseline the same number of attempts, but it uses the same patients, and the validation partition served both for checkpoint and for configuration selection. A confirmatory study should fix one configuration, or a small pre-registered set, and test it against a batch-matched baseline on an untouched cohort. Choosing among configurations that trade accuracy against training and deployment cost is itself a multi-objective search problem; multi-objective evolutionary search~\cite{shirini2024multiobjective} and metaheuristic time--cost balancing~\cite{taherihajivand2024balancing} are possible tools, but they would need to run inside such a nested protocol to avoid selection optimism.

\textbf{Training budget.} In the original comparisons the supervised baseline was trained with batch size 512 and JDCNet with 128, giving JDCNet three times as many optimizer updates. Optimization settings such as the learning rate and the number of training epochs can materially change model performance~\cite{shirini2025impact}. We varied only the batch size, which had no detectable effect on supervised training (Section~\ref{sec:batch_matched_reruns}); the learning rate and the number of epochs were held fixed for all methods.

\textbf{Data and generalization.} The evidence comes from one public source, a binary COVID-19 versus non-COVID task, anteroposterior radiographs, and classes drawn from two BIMCV releases; detector type is associated with the label in the training data. Pairs are same-patient but not necessarily same-visit. On an external MIDRC set, all models, including JDCNet, performed at chance (Section~\ref{sec:external_stress}). Candidate remedies, none evaluated here, include multi-center training on paired cohorts from several hospitals with site-held-out evaluation, domain-generalization methods such as spectral adversarial augmentation~\cite{zhang2023when} or acquisition-noise augmentation~\cite{gholinavaz2025robustness}, harmonization or stratification of detector types, unsupervised domain adaptation, and recalibration of the decision threshold on a small site-specific validation set.

\textbf{Task scope.} The study addresses binary classification because the only public same-patient CT--X-ray cohort with labels available to us is BIMCV-COVID19. The objective extends formally to multi-class and multi-label settings (Section~\ref{sec:theoretical_rationale}), but such extensions require paired multi-disease cohorts with harmonized labels, for example covering pneumonia, tuberculosis, edema, and no-finding categories~\cite{holste2024cxrlt,tiu2022expert}, and are untested.

\textbf{Mechanism.} The teachers are trained on the same patients whose labels they then supervise, so the gate admits most training samples with labels that agree with the ground truth, and the auxiliary target acts mainly as CT-informed sample weighting (Section~\ref{sec:jdcnet_results_main}). Whether cross-fitted teachers, class-conditional thresholds, or calibration methods beyond a single temperature improve on this remains to be tested, as does a factorial comparison with logit KD that varies one design factor at a time.

\textbf{Reporting and statistics.} The 15 fold/seed cells are not independent; the patient-level bootstrap gives wider intervals than the cell-level bootstrap and should be preferred. Comparators received a smaller tuning budget than JDCNet.

\section{Discussion}
\label{sec:discussion}

\textbf{What the evidence supports.} Across 16 JDCNet configurations on the same 510 patients, the paired BA difference against the supervised baseline was positive in 15, with an average near $+0.02$, but no single configuration survived multiplicity adjustment. The pre-registered supplementary experiment changes the reading of this pattern in two ways. Giving the supervised baseline three times as many optimizer updates did not improve it, so the training budget is unlikely to explain the differences. When both methods received the same number of attempts and selected among them on the validation partition, JDCNet retained an advantage of $+0.037$ BA (patient-level CI $[+0.007,+0.068]$). The most defensible conclusion is therefore that confidence-gated CT supervision gives a small within-cohort advantage when its configuration is chosen on validation data, whereas the performance of any particular configuration remains uncertain and the advantage did not transfer to an external set.

\textbf{How the channel might work.} The teacher diagnostics give a concrete picture. Because each teacher was trained on the training partition, it labels those patients with about 95\% accuracy, and the gate admits 84--95\% of them; the admitted argmax targets agree with the ground truth for about 97\%. The auxiliary loss therefore adds weight to patients whose CT is unambiguous and, for the soft-KL variant, conveys the teacher's confidence. JDCNet's shift of the operating point toward higher specificity mirrors the CT-only references, which suggests that the student learns where the CT teacher is confident rather than new labels. Plain and gated logit KD at $T{=}4$ were neutral or harmful, and embedding-level contrastive alignment was near zero. One explanation is that a strongly softened CT distribution or an embedding-alignment objective transfers CT-specific structure that conflicts with the X-ray feature space; because the comparators differ from JDCNet in several factors at once, this remains a hypothesis.

\textbf{Backbone capacity.} On the 226-patient subset, a single BiomedCLIP ViT-B/16 configuration showed no detectable difference from ResNet-18 ($+0.002$, CI $[-0.048,+0.050]$). This comparison is too limited to locate the performance bottleneck, but it gives no indication that backbone capacity alone limits performance in this setting.

\textbf{Implications for evaluation practice.} Three lessons apply beyond this study. First, a positive point estimate with a CI crossing zero on a small cohort is a hypothesis: the balanced-subset gated-logit configuration ($+0.034$) was not reproduced at 510 patients. Second, run-to-run variability can exceed the effect of interest: seeds alone moved supervised BA by $0.041$ (within-cell SD), so best-of-many results need a control with the same number of attempts. Third, reporting the full search with its order, multiplicity-adjusted $p$-values, and a pre-registered nested comparison changes how a result should be read; here it showed that the two configurations that ``passed'' a rule were selected estimates, and replaced them with a method-level comparison whose interval accounts for the selection.

\textbf{Clinical relevance.} JDCNet changes only the training objective, so the deployed classifier takes the same X-ray input and has the same computation graph as a supervised X-ray model. This removes the need for CT at inference, but it does not make the model clinically usable. No triage or prioritization workflow was evaluated, and a higher retrospective BA does not demonstrate better triage. External generalization failed, calibration and thresholds would need site-specific validation, and any clinical use would require prospective evaluation against patient outcomes and assessment under the applicable regulatory framework. Table~\ref{tab:jdcnet_tradeoffs} summarizes benefits and limitations.

\begin{table}[t]
\caption{Benefits and limitations of JDCNet as supported by the present evidence.}
\label{tab:jdcnet_tradeoffs}
\centering
\footnotesize
\setlength{\tabcolsep}{3pt}
\renewcommand{\arraystretch}{0.95}
\begin{tabularx}{\columnwidth}{|>{\raggedright\arraybackslash}p{0.20\columnwidth}|>{\raggedright\arraybackslash}p{0.37\columnwidth}|>{\raggedright\arraybackslash}X|}
\hline
Dimension & Benefit or evidence & Limitation or risk \\ \hline
Deployed model & CT is used only in training; the deployed graph is the supervised ResNet-18 classifier, so per-image inference cost is unchanged. & Same-patient CT is needed for training; one extra teacher forward pass per training sample. \\ \hline
Within-cohort performance & Nested, batch-matched comparison: $+0.037$ BA (patient-level CI $[+0.007,+0.068]$); 15 of 16 configurations positive. & One cohort; no single configuration significant after adjustment; validation partition used for two selections. \\ \hline
Operating point & Higher specificity and AUC, toward the CT teacher's profile. & Sensitivity unchanged or lower; retrospective metrics only. \\ \hline
Generalization & -- & All models at chance on an external radiograph set; single source; binary task. \\ \hline
Clinical translation & X-ray-only inputs avoid CT at inference. & No workflow, outcome, calibration, or regulatory evaluation. \\ \hline
\end{tabularx}
\end{table}

\section{Conclusion}
This paper studied training-only CT supervision for an X-ray classifier whose deployed computation graph is the same as that of the supervised baseline, so it adds no model-inference cost. JDCNet applies a CT-teacher auxiliary target only where the teacher is confident. On 510 same-patient BIMCV patients, 15 of 16 JDCNet configurations had a positive mean difference in balanced accuracy against the supervised baseline, with an average of about $+0.02$, but no single configuration remained significant after multiplicity adjustment. A pre-registered experiment with 240 batch-matched supervised restarts showed that the training budget does not explain these differences and that, when both methods selected one of 16 runs on the validation partition, JDCNet retained an advantage of $+0.037$ BA. Teacher diagnostics indicate that the gate acts mainly as CT-informed sample weighting. Logit distillation and contrastive alignment did not help, and all models, including JDCNet, failed on an external radiograph set. Confidence-gated CT supervision is therefore a small, method-level within-cohort improvement for one public cohort and a binary task. Its value beyond this cohort requires a pre-registered replication on an independent same-patient paired cohort, ideally multi-center and multi-disease.

\section*{Declarations}

\bmhead{Funding}
This research received no specific grant from any funding agency in the public, commercial, or not-for-profit sectors.

\bmhead{Competing interests}
W.Q.Y. serves as a Section Editor for Discover Artificial Intelligence. The
other authors declare no competing interests.

\bmhead{Ethical approval}
This study is a secondary analysis of publicly released, de-identified human
imaging data. The BIMCV-COVID19 data~\cite{delaiglesiavaya2020bimcv}
were collected under approval of the Institutional Review Board of Miguel Hernandez University
and of the local institutional ethics committee at Arnau de Vilanova Hospital,
Valencia Region (CEIm: 12/2020). No new imaging or clinical data were
collected for the present study, so no additional institutional ethical
approval was required. The analysis was conducted under the BIMCV-COVID19 Research Use Agreement,
which requires research-only handling and prohibits re-identification and data
sharing.

\bmhead{Consent to participate}
Not applicable. No patients were recruited or contacted for this
secondary analysis; all imaging data had been de-identified and publicly
released by the data custodians, under the
ethics approval and consent procedures of the original data collection. No
re-identification of individuals was attempted.

\bmhead{Consent to publish}
Not applicable. The manuscript contains no patient images and no identifiable
personal data.

\bmhead{Data availability}
The BIMCV-COVID19 imaging data used in this study are available from the
data custodians (Valencian Region Medical ImageBank) at
\url{https://bimcv.cipf.es/bimcv-projects/bimcv-covid19} under the BIMCV-COVID19
Research Use Agreement at
\url{https://bimcv.cipf.es/bimcv-projects/bimcv-covid19/bimcv-covid19-dataset-research-use-agreement-2/}.
The agreement prohibits data sharing,
re-identification, and publication or reproduction of dataset portions
without permission; no individual image files or dataset are
redistributed. The historical paired subset is part of the public COVID-19
image data collection~\cite{cohen2020covid}, and the external radiographs are
available from MIDRC (\url{https://www.midrc.org}) under its data use terms. The patient-level split definitions, archived aggregate result records, and analysis scripts that support the findings of this study are available at
\url{https://github.com/mabo1215/JDCNET}. Per-image model predictions, which are needed for the patient-level, external, and teacher analyses and are indexed by BIMCV and MIDRC identifiers, are available from the corresponding author on reasonable request, subject to the source data-use terms.

\bmhead{Materials availability}
Not applicable.

\bmhead{Code availability}
Source code, training configurations, and analysis utilities are publicly
available at \url{https://github.com/mabo1215/JDCNET}.

\bmhead{Author contribution}
B.M. conceived the study, developed the method, implemented the experiments,
performed the analysis, and drafted the manuscript. W.Q.Y., J.W., and K.L.
supervised the work. All authors reviewed and approved the final manuscript.

\bibliography{ref}

\begin{thebibliography}{10}
\providecommand{\doi}[1]{\url{https://doi.org/#1}}
\bibcommenthead

\bibitem[\protect\citeauthoryear{Gao et~al.}{2021}]{gao2021covid}
Gao X, Qian Y, Gao A.
\newblock {COVID-ViT}: Classification of {COVID-19} from {CT} Chest Images
  Based on Vision Transformer Models.
\newblock arXiv preprint. 2021;\doi{10.48550/arXiv.2107.01682}.

\bibitem[\protect\citeauthoryear{Saeedi et~al.}{2026}]{saeedi2026depression}
Saeedi N, Shirini K, Gharehveran SS, Rajebi S, Pedrammehr S, Alizadehsani R,
  et~al.
\newblock Depression Detection Using Cross-Attention Multimodal Fusion on
  Audio-Visual Data.
\newblock In: Artificial Intelligence for Neuroscience, Mental Health, and
  Neurodegenerative Disorders ({IWINAC} 2026). vol. 16574 of Lecture Notes in
  Computer Science. Springer, Cham; 2026. p. 243--252.

\bibitem[\protect\citeauthoryear{Raghu et~al.}{2019}]{raghu2019transfusion}
Raghu M, Zhang C, Kleinberg J, Bengio S.
\newblock Transfusion: Understanding Transfer Learning for Medical Imaging.
\newblock In: Advances in Neural Information Processing Systems. vol.~32; 2019.
  .

\bibitem[\protect\citeauthoryear{Zech et~al.}{2018}]{zech2018variable}
Zech JR, Badgeley MA, Liu M, Costa AB, Titano JJ, Oermann EK.
\newblock Variable Generalization Performance of a Deep Learning Model to
  Detect Pneumonia in Chest Radiographs: A Cross-Sectional Study.
\newblock PLoS Medicine. 2018;15(11):e1002683.
\newblock \doi{10.1371/journal.pmed.1002683}.

\bibitem[\protect\citeauthoryear{DeGrave et~al.}{2021}]{degrave2021shortcut}
DeGrave AJ, Janizek JD, Lee SI.
\newblock {AI} for Radiographic {COVID-19} Detection Selects Shortcuts Over
  Signal.
\newblock Nature Machine Intelligence. 2021;3:610--619.
\newblock \doi{10.1038/s42256-021-00338-7}.

\bibitem[\protect\citeauthoryear{Ong~Ly et~al.}{2024}]{ongly2024shortcut}
Ong~Ly C, Unnikrishnan B, Tadic T, Patel T, Duhamel J, Kandel S, et~al.
\newblock Shortcut Learning in Medical {AI} Hinders Generalization: Method for
  Estimating {AI} Model Generalization without External Data.
\newblock npj Digital Medicine. 2024;7(1):124.
\newblock \doi{10.1038/s41746-024-01118-4}.

\bibitem[\protect\citeauthoryear{Sandler et~al.}{2018}]{sandler2018mobilenetv2}
Sandler M, Howard A, Zhu M, Zhmoginov A, Chen LC.
\newblock {MobileNetV2}: Inverted Residuals and Linear Bottlenecks.
\newblock In: Proceedings of the IEEE/CVF Conference on Computer Vision and
  Pattern Recognition (CVPR); 2018. p. 4510--4520.

\bibitem[\protect\citeauthoryear{Tan and Le}{2019}]{tan2019efficientnet}
Tan M, Le QV.
\newblock {EfficientNet}: Rethinking Model Scaling for Convolutional Neural
  Networks.
\newblock In: Proceedings of the International Conference on Machine Learning
  (ICML); 2019. p. 6105--6114.

\bibitem[\protect\citeauthoryear{Hinton et~al.}{2015}]{hinton2015distilling}
Hinton G, Vinyals O, Dean J.
\newblock Distilling the Knowledge in a Neural Network.
\newblock arXiv preprint. 2015;\doi{10.48550/arXiv.1503.02531}.

\bibitem[\protect\citeauthoryear{Romero et~al.}{2015}]{romero2014fitnets}
Romero A, Ballas N, Kahou SE, Chassang A, Gatta C, Bengio Y.
\newblock {FitNets}: Hints for Thin Deep Nets.
\newblock In: International Conference on Learning Representations (ICLR);
  2015. Available from: \url{https://arxiv.org/abs/1412.6550}.

\bibitem[\protect\citeauthoryear{Termritthikun
  et~al.}{2024}]{termritthikun2024explainable}
Termritthikun C, Umer A, Suwanwimolkul S, Xia F, Lee I.
\newblock Explainable Knowledge Distillation for On-Device Chest {X}-Ray
  Classification.
\newblock {IEEE/ACM} Transactions on Computational Biology and Bioinformatics.
  2024;21(4):846--856.
\newblock \doi{10.1109/TCBB.2023.3272333}.

\bibitem[\protect\citeauthoryear{Rajpurkar et~al.}{2017}]{rajpurkar2017chexnet}
Rajpurkar J, Irvin J, Zhu K, Yang B, Mehta H, Duan T, et~al.
\newblock {CheXNet}: Radiologist-Level Pneumonia Detection on Chest {X}-Rays
  with Deep Learning.
\newblock arXiv preprint. 2017;\doi{10.48550/arXiv.1711.05225}.

\bibitem[\protect\citeauthoryear{Irvin et~al.}{2019}]{irvin2019chexpert}
Irvin J, Rajpurkar P, Ko M, Yu Y, Ciurea-Ilcus S, Chute C, et~al.
\newblock {CheXpert}: A Large Chest Radiograph Dataset With Uncertainty Labels
  and Expert Comparison.
\newblock Proceedings of the AAAI Conference on Artificial Intelligence.
  2019;33(1):590--597.
\newblock \doi{10.1609/aaai.v33i01.3301590}.

\bibitem[\protect\citeauthoryear{Johnson et~al.}{2019}]{johnson2019mimiccxr}
Johnson AEW, Pollard TJ, Berkowitz SJ, Greenbaum NR, Lungren MP, Deng Cy,
  et~al.
\newblock {MIMIC-CXR}, a De-Identified Publicly Available Database of Chest
  Radiographs With Free-Text Reports.
\newblock Scientific Data. 2019;6:317.
\newblock \doi{10.1038/s41597-019-0322-0}.

\bibitem[\protect\citeauthoryear{Holste et~al.}{2024}]{holste2024cxrlt}
Holste G, Zhou Y, Wang S, Jaiswal A, Lin M, Zhuge S, et~al.
\newblock Towards Long-Tailed, Multi-Label Disease Classification from Chest
  {X}-Ray: Overview of the {CXR-LT} Challenge.
\newblock Medical Image Analysis. 2024;97:103224.
\newblock \doi{10.1016/j.media.2024.103224}.

\bibitem[\protect\citeauthoryear{Tiu et~al.}{2022}]{tiu2022expert}
Tiu E, Talius E, Patel P, Langlotz CP, Ng AY, Rajpurkar P.
\newblock Expert-Level Detection of Pathologies from Unannotated Chest {X}-Ray
  Images via Self-Supervised Learning.
\newblock Nature Biomedical Engineering. 2022;6(12):1399--1406.
\newblock \doi{10.1038/s41551-022-00936-9}.

\bibitem[\protect\citeauthoryear{Sun et~al.}{2022}]{sun2022performance}
Sun J, Peng L, Li T, Adila D, Zaiman Z, Melton-Meaux GB, et~al.
\newblock Performance of a Chest Radiograph {AI} Diagnostic Tool for
  {COVID-19}: A Prospective Observational Study.
\newblock Radiology: Artificial Intelligence. 2022;4(4):e210217.
\newblock \doi{10.1148/ryai.210217}.

\bibitem[\protect\citeauthoryear{Dosovitskiy et~al.}{2021}]{dosovitskiy2021vit}
Dosovitskiy A, Beyer L, Kolesnikov A, Weissenborn D, Zhai X, Unterthiner T,
  et~al.
\newblock An Image is Worth 16x16 Words: Transformers for Image Recognition at
  Scale.
\newblock In: International Conference on Learning Representations (ICLR);
  2021. .

\bibitem[\protect\citeauthoryear{Liu et~al.}{2021}]{liu2021swin}
Liu Z, Lin Y, Cao Y, Hu H, Wei Y, Zhang Z, et~al.
\newblock Swin Transformer: Hierarchical Vision Transformer using Shifted
  Windows.
\newblock In: Proceedings of the IEEE/CVF International Conference on Computer
  Vision (ICCV); 2021. p. 10012--10022.

\bibitem[\protect\citeauthoryear{Nemati et~al.}{2025}]{nemati2025ferha}
Nemati R, Shirini K, Gharehveran SS.
\newblock {FER-HA}: A Hybrid Attention Model for Facial Emotion Recognition.
\newblock The Journal of Supercomputing. 2025;81(16):1485.
\newblock \doi{10.1007/s11227-025-07983-4}.

\bibitem[\protect\citeauthoryear{Ho and Gwak}{2020}]{ho2020utilizing}
Ho TKK, Gwak J.
\newblock Utilizing Knowledge Distillation in Deep Learning for Classification
  of Chest {X}-Ray Abnormalities.
\newblock IEEE Access. 2020;8:160749--160761.
\newblock \doi{10.1109/ACCESS.2020.3020802}.

\bibitem[\protect\citeauthoryear{Li et~al.}{2020}]{li2020soft}
Li G, Togo R, Ogawa T, Haseyama M.
\newblock Soft-Label Anonymous Gastric {X}-Ray Image Distillation.
\newblock In: 2020 IEEE International Conference on Image Processing (ICIP);
  2020. p. 305--309.

\bibitem[\protect\citeauthoryear{van Sonsbeek
  et~al.}{2021}]{sonsbeek2021variational}
van Sonsbeek T, Zhen X, Worring M, Shao L.
\newblock Variational Knowledge Distillation for Disease Classification in
  Chest {X}-Rays.
\newblock In: Information Processing in Medical Imaging; 2021. p. 334--345.

\bibitem[\protect\citeauthoryear{Yue et~al.}{2021}]{yue2021self}
Yue J, Fang L, Rahmani H, Ghamisi P.
\newblock Self-Supervised Learning with Adaptive Distillation for Hyperspectral
  Image Classification.
\newblock IEEE Transactions on Geoscience and Remote Sensing. 2021;60:1--13.
\newblock \doi{10.1109/TGRS.2021.3057768}.

\bibitem[\protect\citeauthoryear{Rui et~al.}{2024}]{rui2024multimodal}
Rui S, Chen L, Tang Z, Wang L, Liu M, Zhang S, et~al.
\newblock Multi-modal Vision Pre-training for Medical Image Analysis.
\newblock arXiv preprint. 2024;\doi{10.48550/arXiv.2410.10604}.

\bibitem[\protect\citeauthoryear{Zhang et~al.}{2024}]{zhang2023biomedclip}
Zhang S, Xu Y, Usuyama N, Bagga J, Tinn R, Preston S, et~al.
\newblock {BiomedCLIP}: A Multimodal Biomedical Foundation Model Pretrained
  from Fifteen Million Scientific Image-Text Pairs.
\newblock Nature Methods. 2024;21:1560--1570.
\newblock \doi{10.1038/s41592-024-02298-3}.

\bibitem[\protect\citeauthoryear{Vapnik and Vashist}{2009}]{vapnik2009lupi}
Vapnik V, Vashist A.
\newblock A New Learning Paradigm: Learning Using Privileged Information.
\newblock Neural Networks. 2009;22(5--6):544--557.
\newblock \doi{10.1016/j.neunet.2009.06.042}.

\bibitem[\protect\citeauthoryear{Lopez-Paz
  et~al.}{2016}]{lopezpaz2016privileged}
Lopez-Paz D, Bottou L, Sch{\"o}lkopf B, Vapnik V.
\newblock Unifying Distillation and Privileged Information.
\newblock In: International Conference on Learning Representations (ICLR);
  2016. Available from: \url{http://leon.bottou.org/papers/lopez-paz-2016}.

\bibitem[\protect\citeauthoryear{Hoffman
  et~al.}{2016}]{hoffman2016hallucination}
Hoffman J, Gupta S, Darrell T.
\newblock Learning With Side Information Through Modality Hallucination for
  Action Recognition.
\newblock In: Proceedings of the IEEE Conference on Computer Vision and Pattern
  Recognition (CVPR); 2016. p. 826--834.

\bibitem[\protect\citeauthoryear{Chen et~al.}{2022}]{chen2022privileged}
Chen C, Dou Q, Jin Y, Liu Q, Heng PA.
\newblock Learning With Privileged Multimodal Knowledge for Unimodal
  Segmentation.
\newblock {IEEE} Transactions on Medical Imaging. 2022;41(3):621--632.
\newblock \doi{10.1109/TMI.2021.3119385}.

\bibitem[\protect\citeauthoryear{Wang et~al.}{2023}]{wang2023learnable}
Wang H, Ma C, Zhang J, Zhang Y, Avery J, Hull L, et~al.
\newblock Learnable Cross-Modal Knowledge Distillation for Multi-Modal Learning
  with Missing Modality.
\newblock In: Medical Image Computing and Computer Assisted Intervention --
  {MICCAI} 2023. vol. 14223 of Lecture Notes in Computer Science. Springer,
  Cham; 2023. p. 216--226.

\bibitem[\protect\citeauthoryear{Gupta et~al.}{2016}]{gupta2016crossmodal}
Gupta S, Hoffman J, Malik J.
\newblock Cross Modal Distillation for Supervision Transfer.
\newblock In: Proceedings of the IEEE Conference on Computer Vision and Pattern
  Recognition (CVPR); 2016. p. 2827--2836.

\bibitem[\protect\citeauthoryear{Zagoruyko and
  Komodakis}{2017}]{zagoruyko2017attention}
Zagoruyko S, Komodakis N.
\newblock Paying More Attention to Attention: Improving the Performance of
  Convolutional Neural Networks via Attention Transfer.
\newblock In: International Conference on Learning Representations (ICLR);
  2017. .

\bibitem[\protect\citeauthoryear{Tian et~al.}{2020}]{tian2020contrastive}
Tian Y, Krishnan D, Isola P.
\newblock Contrastive Representation Distillation.
\newblock In: International Conference on Learning Representations (ICLR);
  2020. Available from: \url{https://openreview.net/forum?id=SkgpBJrtvS}.

\bibitem[\protect\citeauthoryear{Lin et~al.}{2017}]{lin2017fpn}
Lin TY, Doll{\'a}r P, Girshick R, He K, Hariharan B, Belongie S.
\newblock Feature Pyramid Networks for Object Detection.
\newblock In: Proceedings of the IEEE Conference on Computer Vision and Pattern
  Recognition (CVPR); 2017. p. 2117--2125.

\bibitem[\protect\citeauthoryear{Sarkar and Etemad}{2022}]{sarkar2022xkd}
Sarkar P, Etemad A.
\newblock {XKD}: Cross-Modal Knowledge Distillation with Domain Alignment for
  Video Representation Learning.
\newblock arXiv preprint. 2022;\doi{10.48550/arXiv.2211.13929}.

\bibitem[\protect\citeauthoryear{Ferrod et~al.}{2025}]{ferrod2025crodinokd}
Ferrod R, Dantas CF, Di~Caro L, Ienco D.
\newblock Revisiting Cross-Modal Knowledge Distillation: A Disentanglement
  Approach for {RGBD} Semantic Segmentation.
\newblock arXiv preprint. 2025;\doi{10.48550/arXiv.2505.24361}.

\bibitem[\protect\citeauthoryear{Li et~al.}{2025}]{li2025mstdistill}
Li H, Yang P, Chen J, Dong L, Chen Y, Wang Q.
\newblock {MST-Distill}: Mixture of Specialized Teachers for Cross-Modal
  Knowledge Distillation.
\newblock arXiv preprint. 2025;\doi{10.48550/arXiv.2507.07015}.

\bibitem[\protect\citeauthoryear{Wu et~al.}{2023}]{wu2023krd}
Wu L, Lin H, Huang Y, Li SZ.
\newblock Quantifying the Knowledge in {GNN}s for Reliable Distillation into
  {MLP}s.
\newblock arXiv preprint. 2023;\doi{10.48550/arXiv.2306.05628}.

\bibitem[\protect\citeauthoryear{Zhang et~al.}{2022}]{zhang2022confidence}
Zhang H, Chen D, Wang C.
\newblock Confidence-Aware Multi-Teacher Knowledge Distillation.
\newblock In: {ICASSP} 2022 -- 2022 {IEEE} International Conference on
  Acoustics, Speech and Signal Processing ({ICASSP}); 2022. p. 4498--4502.

\bibitem[\protect\citeauthoryear{Amara et~al.}{2022}]{amara2022bdkd}
Amara I, Sepahvand N, Meyer BH, Gross WJ, Clark JJ.
\newblock {BD-KD}: Balancing the Divergences for Online Knowledge Distillation.
\newblock arXiv preprint. 2022;\doi{10.48550/arXiv.2212.12965}.

\bibitem[\protect\citeauthoryear{Zhao et~al.}{2022}]{zhao2022decoupled}
Zhao B, Cui Q, Song R, Qiu Y, Liang J.
\newblock Decoupled Knowledge Distillation.
\newblock In: 2022 {IEEE/CVF} Conference on Computer Vision and Pattern
  Recognition ({CVPR}); 2022. p. 11943--11952.

\bibitem[\protect\citeauthoryear{Li et~al.}{2023}]{li2023curriculum}
Li Z, Li X, Yang L, Zhao B, Song R, Luo L, et~al.
\newblock Curriculum Temperature for Knowledge Distillation.
\newblock In: Proceedings of the {AAAI} Conference on Artificial Intelligence.
  vol.~37; 2023. p. 1504--1512.

\bibitem[\protect\citeauthoryear{Guo et~al.}{2017}]{guo2017calibration}
Guo C, Pleiss G, Sun Y, Weinberger KQ.
\newblock On Calibration of Modern Neural Networks.
\newblock In: Proceedings of the 34th International Conference on Machine
  Learning. vol.~70 of Proceedings of Machine Learning Research. PMLR; 2017. p.
  1321--1330.
\newblock Available from: \url{https://proceedings.mlr.press/v70/guo17a.html}.

\bibitem[\protect\citeauthoryear{Rajaraman
  et~al.}{2022}]{rajaraman2022calibration}
Rajaraman S, Ganesan P, Antani S.
\newblock Deep Learning Model Calibration for Improving Performance in
  Class-Imbalanced Medical Image Classification Tasks.
\newblock {PLOS ONE}. 2022;17(1):e0262838.
\newblock \doi{10.1371/journal.pone.0262838}.

\bibitem[\protect\citeauthoryear{Cahan et~al.}{2025}]{cahan2025cxrctpa}
Cahan N, Klang E, Greenspan H.
\newblock Cross-Modal {CXR}-{CTPA} Knowledge Distillation Using Latent
  Diffusion Priors.
\newblock In: Medical Image Computing and Computer-Assisted Intervention
  (MICCAI); 2025. .

\bibitem[\protect\citeauthoryear{Zhang et~al.}{2026}]{zhang2026dante}
Zhang Y, Wang H, Liu Z, Sun M.
\newblock Disentangling for Transfer: A Modality-Decoupled Framework for
  Cross-Modal Knowledge Distillation.
\newblock In: Proceedings of the AAAI Conference on Artificial Intelligence
  (AAAI); 2026. .

\bibitem[\protect\citeauthoryear{Zeng and Albarqouni}{2026}]{zeng2026kmat}
Zeng Y, Albarqouni S.
\newblock {K-MaT}: Knowledge Distillation Across Multi-Anatomy Tasks for
  Medical Image Analysis.
\newblock arXiv preprint. 2026;\doi{10.48550/arXiv.2603.06340}.

\bibitem[\protect\citeauthoryear{Kelly et~al.}{2019}]{kelly2019clinicalimpact}
Kelly CJ, Karthikesalingam A, Suleyman M, Corrado G, King D.
\newblock Key Challenges for Delivering Clinical Impact with Artificial
  Intelligence.
\newblock BMC Medicine. 2019;17(1):195.
\newblock \doi{10.1186/s12916-019-1426-2}.

\bibitem[\protect\citeauthoryear{Wynants et~al.}{2020}]{wynants2020prediction}
Wynants L, Van~Calster B, Collins GS, Riley RD, Heinze G, Schuit E, et~al.
\newblock Prediction Models for Diagnosis and Prognosis of {COVID-19}:
  Systematic Review and Critical Appraisal.
\newblock BMJ. 2020;369:m1328.
\newblock \doi{10.1136/bmj.m1328}.

\bibitem[\protect\citeauthoryear{Roberts et~al.}{2021}]{roberts2021pitfalls}
Roberts M, Driggs D, Thorpe M, Gilbey J, Yeung M, Ursprung S, et~al.
\newblock Common Pitfalls and Recommendations for Using Machine Learning to
  Detect and Prognosticate for {COVID}-19 Using Chest Radiographs and {CT}
  Scans.
\newblock Nature Machine Intelligence. 2021;3:199--217.
\newblock \doi{10.1038/s42256-021-00307-0}.

\bibitem[\protect\citeauthoryear{Oakden-Rayner
  et~al.}{2020}]{oakdenrayner2020hidden}
Oakden-Rayner L, Dunnmon J, Carneiro G, Re C.
\newblock Hidden Stratification Causes Clinically Meaningful Failures in
  Machine Learning for Medical Imaging.
\newblock Proceedings of the ACM Conference on Health, Inference, and Learning.
  2020;p. 151--159.
\newblock \doi{10.1145/3368555.3384468}.

\bibitem[\protect\citeauthoryear{Ahluwalia
  et~al.}{2023}]{ahluwalia2023subgroup}
Ahluwalia M, Abdalla M, Sanayei J, Seyyed-Kalantari L, Hussain M, Ali A, et~al.
\newblock The Subgroup Imperative: Chest Radiograph Classifier Generalization
  Gaps in Patient, Setting, and Pathology Subgroups.
\newblock Radiology: Artificial Intelligence. 2023;5(5):e220270.
\newblock \doi{10.1148/ryai.220270}.

\bibitem[\protect\citeauthoryear{Maier-Hein
  et~al.}{2018}]{maierhein2018rankings}
Maier-Hein L, Eisenmann M, Reinke A, Onogur S, Stankovic M, Scholz P, et~al.
\newblock Why Rankings of Biomedical Image Analysis Competitions Should Be
  Interpreted with Care.
\newblock Nature Communications. 2018;9(1):5217.
\newblock \doi{10.1038/s41467-018-07619-7}.

\bibitem[\protect\citeauthoryear{Varoquaux and
  Cheplygina}{2022}]{varoquaux2022failures}
Varoquaux G, Cheplygina V.
\newblock Machine Learning for Medical Imaging: Methodological Failures and
  Recommendations for the Future.
\newblock npj Digital Medicine. 2022;5(1):48.
\newblock \doi{10.1038/s41746-022-00592-y}.

\bibitem[\protect\citeauthoryear{Zhang et~al.}{2023}]{zhang2023when}
Zhang J, Chao H, Dhurandhar A, Chen PY, Tajer A, Xu Y, et~al.
\newblock When Neural Networks Fail to Generalize? {A} Model Sensitivity
  Perspective.
\newblock In: Proceedings of the {AAAI} Conference on Artificial Intelligence.
  vol.~37; 2023. p. 11219--11227.

\bibitem[\protect\citeauthoryear{de~la Iglesia~Vay\'{a}
  et~al.}{2020}]{delaiglesiavaya2020bimcv}
de~la Iglesia~Vay\'{a} M, Saborit JM, Montell JA, Pertusa A, Bustos A, Cazorla
  M, et~al.
\newblock {BIMCV COVID-19+}: a large annotated dataset of {RX} and {CT} images
  from {COVID-19} patients.
\newblock arXiv preprint. 2020;\doi{10.48550/arXiv.2006.01174}.

\bibitem[\protect\citeauthoryear{Cohen et~al.}{2020}]{cohen2020covid}
Cohen JP, Morrison P, Dao L, Roth K, Duong TQ, Ghassemi M.
\newblock {COVID-19} Image Data Collection: Prospective Predictions Are the
  Future.
\newblock Machine Learning for Biomedical Imaging.
  2020;\doi{10.59275/j.melba.2020-48g7}.

\bibitem[\protect\citeauthoryear{{MIDRC}}{}]{midrc}
{MIDRC}.: Medical Imaging and Data Resource Center ({MIDRC}).
\newblock Accessed 2026.
\newblock \url{https://www.midrc.org}.

\bibitem[\protect\citeauthoryear{Shirini
  et~al.}{2024}]{shirini2024multiobjective}
Shirini K, Aghdasi HS, Saeedvand S.
\newblock Multi-Objective Aircraft Landing Problem: A Multi-Population Solution
  Based on Non-Dominated Sorting Genetic Algorithm-{II}.
\newblock The Journal of Supercomputing. 2024;80(17):25283--25314.
\newblock \doi{10.1007/s11227-024-06385-2}.

\bibitem[\protect\citeauthoryear{Taheri~Hajivand
  et~al.}{2024}]{taherihajivand2024balancing}
Taheri~Hajivand A, Shirini K, Samadi~Gharehveran S.
\newblock Balancing Time and Cost in Resource-Constrained Project Scheduling
  Using Meta-Heuristic Approach.
\newblock Journal of Agricultural Machinery. 2024;14(2):215--234.
\newblock \doi{10.22067/jam.2023.81735.1157}.

\bibitem[\protect\citeauthoryear{Shirini et~al.}{2025}]{shirini2025impact}
Shirini K, Kordan MB, Gharehveran SS.
\newblock Impact of Learning Rate and Epochs on {LSTM} Model Performance: A
  Study of Chlorophyll-a Concentrations in the {Marmara Sea}.
\newblock The Journal of Supercomputing. 2025;81(1):265.
\newblock \doi{10.1007/s11227-024-06806-2}.

\bibitem[\protect\citeauthoryear{Gholinavaz
  et~al.}{2025}]{gholinavaz2025robustness}
Gholinavaz S, Saeedi N, Gharehveran SS.
\newblock Robustness Analysis of {YOLO} and {Faster R-CNN} for Object Detection
  in Realistic Weather Scenarios with Noise Augmentation.
\newblock Scientific Reports. 2025;15(1):44888.
\newblock \doi{10.1038/s41598-025-28737-5}.

\end{thebibliography}

\clearpage
\begin{appendices}

\section{Supplementary Evidence and Interpretation}

This appendix documents how the patient-level comparisons were computed and read. All entries are support for, or constraints on, the main-text findings rather than independent clinical validation.

\subsection{How to Read the Evidence}

Each comparison is paired: for each held-out fold and seed, the supervised X-ray model and the tested model are evaluated on the same patients, and the tables report the tested model's balanced accuracy (BA) minus the matched supervised BA. BA gives equal weight to sensitivity and specificity, which matters for the imbalanced 510-patient cohort. The decision rule (mean $\Delta\mathrm{BA}\geq+0.03$ and lower 95\% bootstrap CI bound $>0$) is a descriptive screen, not a confirmatory test; because the JDCNet configurations were explored adaptively on the same cohort (main text, Section~\ref{sec:statistical_protocol}), multiplicity-adjusted $p$-values and unselected averages are reported alongside it. CT-only reference rows indicate whether the CT view carries class signal for the same patients; transfer rows indicate whether a training objective moves the X-ray student.

\subsection{Directly Computed Tests for All JDCNet Configurations}
\label{sec:app_initial_grid}

The per fold/seed test-partition records of all 16 configurations are archived, so every test can be computed directly from the 15 paired cells (Table~\ref{tab:app_initial_grid_tests}). The paired $t$-test, Wilcoxon signed-rank test, bootstrap test, and exact sign test lead to the same reading: a few configurations have unadjusted $p<0.05$, none remains significant after Benjamini--Hochberg adjustment over the 16 configurations, and the per-cell average over all 16 is positive. The mean change in specificity is positive for all 16 configurations, whereas the mean change in sensitivity is negative for 12.

\begin{table}[htbp]
\caption{Tests computed directly from the archived per fold/seed records of the 16 JDCNet configurations (510-patient cohort; $n{=}15$ paired cells; differences against the batch-512 supervised baseline). Stage: I, initial; A, extension A; B, extension B. $p_t$: paired $t$-test; $p_W$: Wilcoxon signed-rank; $p_b$: bootstrap (shift method, 10{,}000 resamples); $p_s$: exact sign test; $p_{BH}$: Benjamini--Hochberg-adjusted $p_t$. $\Delta$Sens, $\Delta$Spec, $\Delta$AUC: mean paired differences in sensitivity, specificity, and AUC.}
\label{tab:app_initial_grid_tests}
\centering
\scriptsize
\setlength{\tabcolsep}{1.1pt}
\renewcommand{\arraystretch}{0.9}
\begin{tabular*}{\textwidth}{@{\extracolsep{\fill}}lccccrcccccrrr@{}}
\toprule
View & $\tau$ & $\lambda$ & Target & Stage & $\Delta\mathrm{BA}$ & $p_t$ & $p_W$ & $p_b$ & $p_s$ & $p_{BH}$ & $\Delta$Sens & $\Delta$Spec & $\Delta$AUC \\
\midrule
3-slice & 0.70 & 1.0 & soft & B & $+0.0345$ & 0.014 & 0.022 & 0.003 & 0.30 & 0.22 & $-0.022$ & $+0.091$ & $+0.039$ \\
Mid & 0.80 & 1.5 & hard & A & $+0.0329$ & 0.028 & 0.035 & 0.010 & 0.18 & 0.22 & $+0.013$ & $+0.053$ & $+0.018$ \\
Mid & 0.70 & 1.0 & hard & I & $+0.0298$ & 0.079 & 0.095 & 0.050 & 0.30 & 0.22 & $-0.001$ & $+0.061$ & $+0.033$ \\
Mid & 0.70 & 1.5 & hard & A & $+0.0296$ & 0.061 & 0.14 & 0.032 & 0.61 & 0.22 & $-0.005$ & $+0.064$ & $+0.006$ \\
Mid & 0.70 & 1.0 & soft & B & $+0.0264$ & 0.082 & 0.14 & 0.051 & 0.30 & 0.22 & $+0.016$ & $+0.036$ & $+0.022$ \\
3-slice & 0.80 & 1.0 & soft & B & $+0.0258$ & 0.11 & 0.14 & 0.082 & 0.30 & 0.23 & $-0.021$ & $+0.073$ & $+0.026$ \\
3-slice & 0.80 & 1.0 & hard & I & $+0.0247$ & 0.082 & 0.17 & 0.050 & 0.30 & 0.22 & $-0.037$ & $+0.086$ & $+0.027$ \\
3-slice & 0.70 & 0.5 & hard & I & $+0.0239$ & 0.11 & 0.14 & 0.087 & 0.30 & 0.23 & $-0.005$ & $+0.053$ & $+0.018$ \\
Mid & 0.80 & 1.0 & soft & B & $+0.0231$ & 0.19 & 0.23 & 0.15 & 0.61 & 0.33 & $-0.044$ & $+0.091$ & $+0.018$ \\
3-slice & 0.80 & 1.5 & hard & A & $+0.0181$ & 0.25 & 0.25 & 0.22 & 0.61 & 0.36 & $-0.018$ & $+0.054$ & $+0.038$ \\
3-slice & 0.80 & 0.5 & hard & I & $+0.0151$ & 0.40 & 0.30 & 0.37 & 0.61 & 0.54 & $+0.016$ & $+0.014$ & $+0.018$ \\
Mid & 0.80 & 0.5 & hard & I & $+0.0144$ & 0.22 & 0.23 & 0.18 & 0.30 & 0.34 & $+0.016$ & $+0.013$ & $+0.014$ \\
Mid & 0.80 & 1.0 & hard & I & $+0.0127$ & 0.44 & 0.49 & 0.41 & 0.30 & 0.54 & $-0.019$ & $+0.045$ & $+0.016$ \\
3-slice & 0.70 & 1.0 & hard & I & $+0.0066$ & 0.63 & 0.76 & 0.63 & 1.00 & 0.72 & $-0.052$ & $+0.065$ & $+0.012$ \\
Mid & 0.70 & 0.5 & hard & I & $+0.0005$ & 0.97 & 0.98 & 0.97 & 1.00 & 0.97 & $-0.061$ & $+0.062$ & $+0.006$ \\
3-slice & 0.70 & 1.5 & hard & A & $-0.0031$ & 0.84 & 0.93 & 0.83 & 0.61 & 0.90 & $-0.077$ & $+0.071$ & $0.000$ \\
\multicolumn{5}{@{}l}{Average of the 8 initial} & $+0.0160$ & 0.11 & 0.28 & 0.078 & 1.00 & -- & -- & -- & -- \\
\multicolumn{5}{@{}l}{Average of all 16} & $+0.0197$ & 0.060 & 0.026 & 0.032 & 0.007 & -- & -- & -- & -- \\
\bottomrule
\end{tabular*}
\end{table}

\subsection{Cohort Construction and Leakage Audit}
\label{sec:app_cohort_construction}

The 510-patient cohort was assembled with the following rules. (i)~Patients are taken from the BIMCV-COVID19+ release (label COVID-19) and the BIMCV-COVID19$-$ release (label non-COVID); every patient for whom at least one chest CT volume and at least one chest radiograph could be retrieved is included. No acquisition-date window was imposed between the two studies. (ii)~Each patient contributes exactly one radiograph and one CT volume: the largest CT file (typically the full-volume acquisition) and one radiograph, chosen deterministically as the first after sorting by file path. All selected radiographs are anteroposterior. (iii)~Labels are patient-level and inherited from the release. (iv)~Patients are assigned to five folds by a seeded random permutation within each class. For fold $k$, fold $k$ is the test partition, fold $(k{+}1)\bmod5$ the validation partition, and the remaining three folds the training partition. (v)~Because each patient has one row, no patient's radiograph or CT appears in more than one partition of a split, and all methods are evaluated on identical test patients within each fold and seed. (vi)~CT teachers, supervised baselines, and JDCNet students are trained only on the training partition of each fold and seed, with checkpoint selection on the validation partition; test partitions are used only for the final evaluation. (vii)~Inputs are scaled to $[0,1]$ without dataset-level intensity statistics, so no normalization statistics are shared across partitions.

Rule (i) has two consequences for interpretation. The CT and radiograph of a patient may come from different visits, which weakens the pairing; and because the two classes come from two releases, a classifier may exploit release-specific acquisition cues (main text, Section~\ref{sec:external_stress}). Figure~\ref{fig:app_cohort_flow} summarizes the construction.

\begin{figure}[htbp]
\centering
\resizebox{\linewidth}{!}{%
\begin{tikzpicture}[
  main/.style={draw, rounded corners, align=center, minimum width=66mm, minimum height=8mm, font=\scriptsize, fill=blue!6, draw=blue!55, inner sep=3pt},
  excl/.style={draw, rounded corners, align=center, minimum width=48mm, minimum height=8mm, font=\scriptsize, fill=red!5, draw=red!55, inner sep=3pt},
  final/.style={draw, rounded corners, align=center, minimum width=66mm, minimum height=11mm, font=\scriptsize\bfseries, fill=green!8, draw=green!55, inner sep=3pt},
  arr/.style={-{Latex[length=1.8mm]}, thick}
]
\node[main] (n1) at (0,0) {BIMCV-COVID19+ (COVID-19) and BIMCV-COVID19$-$ (non-COVID)\\public releases};
\node[main] (n2) at (0,-1.7) {Patients with $\geq$1 retrievable chest CT volume \emph{and}\\$\geq$1 retrievable chest radiograph (no date window)};
\node[excl] (e2) at (6.7,-1.7) {Excluded: CT or radiograph\\missing or not retrievable};
\node[main] (n3) at (0,-3.4) {One radiograph and one CT volume per patient\\(largest CT file; first radiograph by path)};
\node[main] (n4) at (0,-5.1) {Patient-level label from the release;\\seeded random fold assignment within each class};
\node[final] (n5) at (0,-7.0) {Final cohort: 510 patients\\(113 COVID-19 / 397 non-COVID); per fold: test $k$,\\validation $(k{+}1)\bmod 5$, training on the remaining three folds};
\draw[arr] (n1) -- (n2);
\draw[arr] (n2) -- (n3);
\draw[arr] (n3) -- (n4);
\draw[arr] (n4) -- (n5);
\draw[arr] (n2.east) -- (e2.west);
\end{tikzpicture}%
}
\caption{Construction of the 510-patient same-patient paired BIMCV cohort, following rules (i)--(iv). Teachers, baselines, and students are trained within the training partition of each fold and seed (rule (vi)).}
\label{fig:app_cohort_flow}
\end{figure}

\subsection{Comparator Interpretation}

Table~\ref{tab:app_comparator_summary} reads the CT-only references and the transfer rows together. The CT-only references show that the mid-slice and 3-slice CT views carry class signal for the same patients. The transfer rows show that, among the tested channels, only the JDCNet configurations are mostly positive, and that their advantage survives a nested, batch-matched comparison within this cohort.

\begin{table}[htbp]
\caption{Comparator interpretation. Unless marked otherwise, rows are paired differences against the supervised X-ray baseline on the 510-patient cohort.}
\label{tab:app_comparator_summary}
\centering
\scriptsize
\setlength{\tabcolsep}{2.5pt}
\renewcommand{\arraystretch}{0.9}
\begin{tabularx}{\columnwidth}{@{}>{\raggedright\arraybackslash}p{0.21\columnwidth}>{\raggedright\arraybackslash}p{0.40\columnwidth}>{\raggedright\arraybackslash}X@{}}
\toprule
Family & Representative result & Reading \\
\midrule
CT-only reference & Mid $+0.045$ [$+0.019,+0.070$]; 3-slice $+0.051$ [$+0.025,+0.079$] & CT carries signal \\
JDCNet & 15/16 configurations positive; nested, batch-matched comparison $+0.037$ [$+0.016,+0.058$] & Small within-cohort gain \\
Gated logit KD & Mid $-0.022$ [$-0.055,+0.010$]; 3-slice $-0.026$; mean proj.\ $-0.014$; DRR $-0.064$ [$-0.095,-0.034$] & No gain \\
Plain logit KD & 3-slice $-0.036$ [$-0.062,-0.013$]; other views $-0.016$ to $-0.001$, CIs include zero & No gain \\
Contrastive alignment & Best of 4: $+0.008$ [$-0.021,+0.038$] & No gain \\
BiomedCLIP (226 subset) & $+0.002$ [$-0.048,+0.050$] vs.\ ResNet-18 on the same subset & No gain \\
Gated logit KD (earlier balanced-subset scan) & Best of eight configurations $+0.034$, CI crosses zero & Not reproduced at 510 \\
\bottomrule
\end{tabularx}
\end{table}

\subsection{Full Contrastive-Alignment Sweep}
\label{sec:app_contrastive_510}

Table~\ref{tab:app_contrastive_510} reports all four contrastive-alignment configurations. Two teacher views (mid-slice and 3-slice, the views whose CT-only references exceed the baseline) and two InfoNCE temperatures ($\tau_{\mathrm{NCE}}\in\{0.07,0.20\}$) were evaluated with the same folds, seeds, ResNet-18 student, and weighted sampler as the JDCNet sweep. The two-stage protocol (1)~trains a paired CT/X-ray encoder pair with projection heads using symmetric InfoNCE on patient-paired batches for 100 epochs, then (2)~discards the CT branch, replaces the X-ray head with a linear classifier, and fine-tunes for 50 epochs with weighted cross-entropy on the same labeled training partition. No configuration shows a gain; the best (3-slice, $\tau_{\mathrm{NCE}}{=}0.20$) yields $+0.008$ with 7 positive, 1 tied, and 7 negative cells, and both mid-slice configurations are negative on the point estimate.

\begin{table}[htbp]
\caption{Contrastive alignment (InfoNCE pretraining followed by supervised fine-tuning) on the 510-patient cohort ($n=15$ paired fold/seed cells per row).}
\label{tab:app_contrastive_510}
\centering
\scriptsize
\setlength{\tabcolsep}{3pt}
\renewcommand{\arraystretch}{0.9}
\begin{tabularx}{\columnwidth}{@{}>{\raggedright\arraybackslash}Xcccc@{}}
\toprule
Teacher view & $\tau_{\mathrm{NCE}}$ & BA mean [95\% CI] & $\Delta\mathrm{BA}$ [95\% CI] & $+/0/-$ \\
\midrule
3-slice stack & 0.20 & 0.633 [0.610, 0.656] & $+0.008$ [$-0.021,+0.038$] & 7/1/7 \\
3-slice stack & 0.07 & 0.627 [0.594, 0.657] & $+0.003$ [$-0.023,+0.030$] & 7/0/8 \\
Mid-slice     & 0.20 & 0.620 [0.594, 0.643] & $-0.005$ [$-0.030,+0.020$] & 5/0/10 \\
Mid-slice     & 0.07 & 0.616 [0.590, 0.642] & $-0.008$ [$-0.038,+0.027$] & 5/0/10 \\
\bottomrule
\end{tabularx}
\end{table}

\subsection{CT View Comparison on the 226-Patient Subset}
\label{sec:app_ct_variants}

Before the 510-patient experiments, the four CT views (mid-slice, 3-slice stack, axial mean projection, DRR) were compared on a class-balanced 226-patient BIMCV subset with five-fold patient-level cross-validation and seeds 42--44 (Table~\ref{tab:app_ct_variants}). Each view was used for a CT-only reference, plain logit KD, and gated logit KD ($T{=}4$, $\alpha{=}0.6$, gate threshold $0.50$ with the teacher-correctness filter of Eq.~(\ref{eq:logit_kd_loss})). On this subset only the mean-projection CT-only reference clearly exceeded the X-ray baseline, and no logit-KD configuration showed a gain for any view. On the 510-patient cohort the mid-slice and 3-slice references exceeded the baseline while the mean projection did not, and the logit-KD comparators remained negative or neutral for all four views.

\begin{table}[htbp]
\caption{CT view comparison on the class-balanced 226-patient BIMCV subset (five folds $\times$ seeds 42--44, $n{=}15$ cells per row). $\Delta\mathrm{BA}$ is the paired difference against the supervised X-ray baseline trained on the same subset.}
\label{tab:app_ct_variants}
\centering
\scriptsize
\setlength{\tabcolsep}{2.5pt}
\renewcommand{\arraystretch}{0.88}
\begin{tabularx}{\columnwidth}{@{}p{0.24\columnwidth}lXc@{}}
\toprule
Teacher view & Method & $\Delta\mathrm{BA}$ [95\% CI] & $+/0/-$ \\
\midrule
Mid-slice            & CT-only ref.  & $+0.011$ [$-0.024,+0.046$] & 8/0/7 \\
                     & Plain KD      & $-0.022$ [$-0.062,+0.021$] & 6/0/9 \\
                     & Gated KD      & $-0.014$ [$-0.053,+0.023$] & 8/0/7 \\
3-slice stack        & CT-only ref.  & $+0.019$ [$-0.013,+0.058$] & 8/3/5 \\
                     & Plain KD      & $-0.030$ [$-0.083,+0.016$] & 7/0/8 \\
                     & Gated KD      & $-0.045$ [$-0.092,+0.000$] & 5/0/10 \\
Mean projection      & CT-only ref.  & $+0.045$ [$+0.008,+0.081$] & 9/1/5 \\
                     & Plain KD      & $-0.031$ [$-0.082,+0.019$] & 5/1/9 \\
                     & Gated KD      & $-0.015$ [$-0.058,+0.027$] & 6/2/7 \\
DRR                  & CT-only ref.  & $+0.023$ [$-0.023,+0.066$] & 9/0/6 \\
                     & Plain KD      & $-0.043$ [$-0.095,+0.014$] & 4/1/10 \\
                     & Gated KD      & $-0.005$ [$-0.048,+0.039$] & 8/0/7 \\
\bottomrule
\end{tabularx}
\end{table}

\subsection{Earlier Gated Logit-KD Scan on a Balanced Subset}
\label{sec:app_calibration_scan}

Before the CT-view comparison of Table~\ref{tab:app_ct_variants}, a scan of gated logit KD over eight combinations of $T\in\{2,4,8\}$ and gate threshold $\in\{0.50,0.55,0.60\}$ ($T{=}4$ with threshold $0.55$ was not run; five folds $\times$ three seeds per configuration) was run on a class-balanced BIMCV subset of similar size, with a different split, teacher run, and training configuration, and against the baseline of that run. Its best configuration ($T{=}4$, threshold $0.50$) reached $\Delta\mathrm{BA}=+0.034$ with a CI crossing zero ($[-0.004,+0.073]$). Because the split and training configuration differ, this value is not directly comparable with the gated-KD rows of Table~\ref{tab:app_ct_variants}, which use the same $T$ and threshold but are negative or neutral for every view. With eight configurations scanned, this best value is itself a selected estimate, and on the 510-patient cohort the gated-logit configurations were negative or neutral. The scan illustrates the selection and small-sample effects that the main text addresses for JDCNet.

\subsection{Earlier Gate-Temperature Reruns on a Different Host}
\label{sec:app_gate_temperature}

Before the pre-registered experiment of Section~\ref{sec:batch_matched_reruns}, the two selected configurations were rerun on a different host with the supervised baseline and JDCNet trained with the same batch size (128) and schedule, varying the gate temperature $T_{\mathrm{g}}\in\{0.5,1.0,2.0\}$. For a two-class teacher this is equivalent to changing the raw-confidence threshold (main text, Eq.~(\ref{eq:temperature_threshold})); the equivalent raw thresholds are listed in Table~\ref{tab:app_gate_temperature}. The rerun used a separate five-fold split of the BIMCV paired pool, restricted to the 466 patients for whom a CT teacher view could be rendered on the rerun host, a single seed ($n{=}5$ folds per row), and reports BA on the validation partition, which was also used for checkpoint selection; values are therefore optimistic for both arms and not comparable with the test-fold values of the main sweep. Both configurations are positive at every temperature, and within-configuration changes are at most $0.025$. The 3-slice soft-KL intervals at $T_{\mathrm{g}}{=}1.0$ and $2.0$ include zero. An earlier rerun of the same design on a class-balanced BIMCV subset (216 patients with a rendered CT teacher view) with three seeds did not reproduce a gain (3-slice soft-KL about $-0.09$ with a CI excluding zero; mid-slice hard about $-0.02$ with a CI including zero; within-configuration changes across temperatures at most $0.013$). The per-run records of both reruns were later recovered from a mirror of that host; JDCNet test-partition metrics exist, but the supervised baselines of these reruns were evaluated only on the validation partition, so the paired comparison is available only on that partition. Both reruns are superseded by the pre-registered experiment. The reruns were prepared on a different host from the main sweep. On that host, CT volumes were not available for some positive patients; for those patients the preprocessing approximated the 3-slice and mean-projection views by replicating a previously extracted mid-slice image, and the number of affected patients was not recorded. The 3-slice teacher of both reruns is therefore partly a mid-slice teacher, which further limits their comparability with the main sweep, where every view was rendered from the CT volume.

\begin{table}[htbp]
\caption{Batch-matched gate-temperature rerun on a 466-patient split (single seed, $n{=}5$ folds per row; validation-partition BA). $\tau'$: raw-confidence threshold equivalent to gating at $\tau$ with temperature $T_{\mathrm{g}}$. $\Delta\mathrm{BA}$ is the paired difference against the batch-matched supervised baseline (validation BA $0.664$ for the 3-slice row set and $0.604$ for the mid-slice row set, which use different student manifests); the last column is the change relative to $T_{\mathrm{g}}{=}1.0$.}
\label{tab:app_gate_temperature}
\centering
\scriptsize
\setlength{\tabcolsep}{3pt}
\renewcommand{\arraystretch}{0.9}
\begin{tabularx}{\columnwidth}{@{}>{\raggedright\arraybackslash}Xccccc@{}}
\toprule
Configuration & $T_{\mathrm{g}}$ & $\tau'$ & BA & $\Delta\mathrm{BA}$ [95\% CI] & vs.\ $T_{\mathrm{g}}{=}1.0$ \\
\midrule
3-slice soft-KL, $\tau{=}0.70$ & 0.5 & 0.60 & 0.711 & $+0.047$ [$+0.028,+0.064$] & $+0.025$ \\
3-slice soft-KL, $\tau{=}0.70$ & 1.0 & 0.70 & 0.686 & $+0.022$ [$-0.011,+0.068$] & $0.000$ \\
3-slice soft-KL, $\tau{=}0.70$ & 2.0 & 0.84 & 0.691 & $+0.027$ [$-0.001,+0.061$] & $+0.005$ \\
\midrule
Mid hard, $\tau{=}0.80$  & 0.5 & 0.67 & 0.680 & $+0.076$ [$+0.035,+0.117$] & $+0.007$ \\
Mid hard, $\tau{=}0.80$  & 1.0 & 0.80 & 0.673 & $+0.069$ [$+0.034,+0.106$] & $0.000$ \\
Mid hard, $\tau{=}0.80$  & 2.0 & 0.94 & 0.659 & $+0.054$ [$+0.024,+0.088$] & $-0.014$ \\
\bottomrule
\end{tabularx}
\end{table}

\subsection{Pre-Registered Supplementary Experiment}
\label{sec:app_supplementary}

The analysis plan for this experiment was written and committed to the code repository before any run. On the original 510-patient splits and with the original training code, 16 supervised restarts were trained in each of the 15 fold/seed cells at batch size 128 (training seeds $s+1000r$ for cell seed $s$ and restart $r=0,\dots,15$; 240 runs); everything else matched the original supervised configuration. Every checkpoint of the original sweep and of the restarts was then re-run to store per-patient probabilities on the test partition and on the external set; the recomputed test BA matched the recorded value for 613 of the 615 runs with recorded test metrics and differed by $0.006$ and $0.022$ BA (one patient) for the other two, one initial-grid JDCNet run and one supervised restart. All cell-level statistics use the recorded values; the patient-level and external analyses use the stored predictions. Table~\ref{tab:app_supplementary} lists the nested and summary comparisons, and Table~\ref{tab:app_vs_r0} lists the batch-matched comparison of every configuration (pre-specified analysis~1) with its patient-level bootstrap (analysis~3). Against $r{=}0$, all 16 configurations have a positive mean difference, and none remains significant after Holm adjustment at either level. The pre-specified external comparison (analysis~4) is reported in Section~\ref{sec:external_stress}: McNemar tests per checkpoint pair gave $p<0.05$ in 2 of 15 pairs against the batch-512 students, both favoring the supervised model, and in none of 15 against the batch-128 students.

Two features of the nested comparison are worth noting. First, validation-based selection chose 11 different JDCNet configurations across the 15 cells, and never the configuration with the best test-partition mean (3-slice soft-KL), so the advantage is a property of the configuration family rather than of one configuration. Second, the selected supervised restarts were spread over 11 different restart indices, and their mean validation BA ($0.731$) was close to that of the selected JDCNet runs ($0.745$), so the supervised baseline was not handicapped in the selection step.

\begin{table}[htbp]
\caption{Pre-registered supplementary comparisons on the 510-patient splits (batch-matched supervised restarts). Cell-level: paired over 15 fold/seed cells (bootstrap CI, paired $t$-test). Patient-level: paired bootstrap over 510 patients with out-of-fold predictions pooled per seed (10{,}000 resamples).}
\label{tab:app_supplementary}
\centering
\scriptsize
\setlength{\tabcolsep}{2.0pt}
\renewcommand{\arraystretch}{0.92}
\begin{tabularx}{\columnwidth}{@{}>{\raggedright\arraybackslash}Xcccc@{}}
\toprule
Comparison & Cell-level $\Delta\mathrm{BA}$ [95\% CI] & $p$ & Patient-level $\Delta\mathrm{BA}$ [95\% CI] & $p$ \\
\midrule
Supervised batch 128 ($r{=}0$) $-$ batch 512 & $-0.005$ [$-0.032$, $+0.022$] & 0.75 & $-0.004$ [$-0.038$, $+0.029$] & 0.82 \\
Supervised best of 16 $-$ supervised $r{=}0$ & $+0.013$ [$-0.011$, $+0.038$] & 0.33 & -- & -- \\
JDCNet best of 16 $-$ supervised best of 16 (primary) & $+0.037$ [$+0.016$, $+0.058$] & 0.005 & $+0.037$ [$+0.007$, $+0.068$] & 0.017 \\
JDCNet best of 8 $-$ supervised best of 8 & $+0.029$ [$+0.007$, $+0.049$] & 0.022 & -- & -- \\
JDCNet best of 16 $-$ supervised batch 512 & $+0.046$ [$+0.021$, $+0.072$] & 0.004 & $+0.046$ [$+0.010$, $+0.082$] & 0.012 \\
3-slice soft-KL $-$ supervised $r{=}0$ & $+0.039$ [$+0.002$, $+0.074$] & 0.063 & $+0.039$ [$+0.003$, $+0.077$] & 0.036 \\
Mid hard $-$ supervised $r{=}0$ & $+0.038$ [$+0.009$, $+0.069$] & 0.035 & $+0.038$ [$+0.001$, $+0.073$] & 0.045 \\
Average of 16 configurations $-$ mean of 16 restarts & $+0.019$ [$+0.006$, $+0.031$] & 0.016 & -- & -- \\
\bottomrule
\end{tabularx}
\end{table}

\begin{table}[htbp]
\caption{Pre-specified batch-matched comparison of every JDCNet configuration against the first batch-128 supervised restart ($r{=}0$) of the same cell (510-patient cohort; rows in the order of Table~\ref{tab:app_initial_grid_tests}, where the exploration stage of each configuration is listed). Cell level: mean paired difference over the 15 fold/seed cells, bootstrap CI, and paired $t$-test $p_t$. Patient level: paired bootstrap over the 510 patients with out-of-fold predictions pooled per seed (10{,}000 resamples) and two-sided bootstrap $p_b$. $p_{\mathrm{H}}$: Holm-adjusted over the 16 configurations.}
\label{tab:app_vs_r0}
\centering
\scriptsize
\setlength{\tabcolsep}{1.0pt}
\renewcommand{\arraystretch}{0.9}
\begin{tabular*}{\textwidth}{@{\extracolsep{\fill}}lcccccccccccc@{}}
\toprule
 & & & & \multicolumn{5}{c}{Cell level} & \multicolumn{3}{c}{Patient level} \\
\cmidrule(lr){5-9}\cmidrule(l){10-12}
View & $\tau$ & $\lambda$ & Target & $+/0/-$ & $\Delta\mathrm{BA}$ & 95\% CI & $p_t$ & $p_{\mathrm{H}}$ & 95\% CI & $p_b$ & $p_{\mathrm{H}}$ \\
\midrule
3-slice & 0.70 & 1.0 & soft & 11/0/4 & $+0.039$ & [$+0.002$, $+0.074$] & 0.063 & 0.95 & [$+0.003$, $+0.077$] & 0.036 & 0.54 \\
Mid & 0.80 & 1.5 & hard & 9/0/6 & $+0.038$ & [$+0.009$, $+0.069$] & 0.035 & 0.56 & [$+0.001$, $+0.073$] & 0.045 & 0.63 \\
Mid & 0.70 & 1.0 & hard & 9/0/6 & $+0.034$ & [$+0.000$, $+0.071$] & 0.086 & 1.00 & [$-0.001$, $+0.071$] & 0.061 & 0.72 \\
Mid & 0.70 & 1.5 & hard & 9/1/5 & $+0.034$ & [$+0.004$, $+0.069$] & 0.064 & 0.95 & [$+0.005$, $+0.064$] & 0.021 & 0.33 \\
Mid & 0.70 & 1.0 & soft & 9/0/6 & $+0.031$ & [$-0.000$, $+0.064$] & 0.091 & 1.00 & [$-0.000$, $+0.062$] & 0.050 & 0.66 \\
3-slice & 0.80 & 1.0 & soft & 10/0/5 & $+0.030$ & [$-0.005$, $+0.065$] & 0.12 & 1.00 & [$-0.002$, $+0.063$] & 0.065 & 0.72 \\
3-slice & 0.80 & 1.0 & hard & 9/1/5 & $+0.029$ & [$-0.006$, $+0.064$] & 0.14 & 1.00 & [$-0.002$, $+0.062$] & 0.069 & 0.72 \\
3-slice & 0.70 & 0.5 & hard & 10/0/5 & $+0.029$ & [$+0.000$, $+0.057$] & 0.077 & 1.00 & [$-0.001$, $+0.060$] & 0.060 & 0.72 \\
Mid & 0.80 & 1.0 & soft & 10/0/5 & $+0.028$ & [$-0.011$, $+0.065$] & 0.19 & 1.00 & [$-0.008$, $+0.065$] & 0.13 & 1.00 \\
3-slice & 0.80 & 1.5 & hard & 9/0/6 & $+0.023$ & [$-0.016$, $+0.064$] & 0.30 & 1.00 & [$-0.017$, $+0.061$] & 0.26 & 1.00 \\
3-slice & 0.80 & 0.5 & hard & 11/0/4 & $+0.020$ & [$-0.003$, $+0.041$] & 0.11 & 1.00 & [$-0.012$, $+0.052$] & 0.20 & 1.00 \\
Mid & 0.80 & 0.5 & hard & 10/0/5 & $+0.019$ & [$-0.011$, $+0.051$] & 0.26 & 1.00 & [$-0.011$, $+0.051$] & 0.21 & 1.00 \\
Mid & 0.80 & 1.0 & hard & 9/0/6 & $+0.017$ & [$-0.016$, $+0.049$] & 0.33 & 1.00 & [$-0.015$, $+0.050$] & 0.29 & 1.00 \\
3-slice & 0.70 & 1.0 & hard & 9/1/5 & $+0.011$ & [$-0.012$, $+0.035$] & 0.39 & 1.00 & [$-0.024$, $+0.046$] & 0.52 & 1.00 \\
Mid & 0.70 & 0.5 & hard & 9/0/6 & $+0.005$ & [$-0.026$, $+0.034$] & 0.75 & 1.00 & [$-0.030$, $+0.040$] & 0.77 & 1.00 \\
3-slice & 0.70 & 1.5 & hard & 8/0/7 & $+0.002$ & [$-0.035$, $+0.039$] & 0.94 & 1.00 & [$-0.036$, $+0.037$] & 0.95 & 1.00 \\
\multicolumn{4}{@{}l}{Average of all 16} & 10/0/5 & $+0.024$ & [$-0.000$, $+0.051$] & 0.092 & -- & -- & -- & -- \\
\bottomrule
\end{tabular*}
\end{table}
\end{appendices}

\end{document}